# Conformant Planning via Symbolic Model Checking

**Alessandro Cimatti**                                              CIMATTI@IRST.ITC.IT
ITC-IRST, *Via Sommarive 18, 38055 Povo, Trento, Italy*

**Marco Roveri**                                                   ROVERI@IRST.ITC.IT
ITC-IRST, *Via Sommarive 18, 38055 Povo, Trento, Italy*
DSI, *University of Milano, Via Comelico 39, 20135 Milano, Italy*

## Abstract

We tackle the problem of planning in nondeterministic domains, by presenting a new approach to conformant planning. Conformant planning is the problem of finding a sequence of actions that is guaranteed to achieve the goal despite the nondeterminism of the domain. Our approach is based on the representation of the planning domain as a finite state automaton. We use Symbolic Model Checking techniques, in particular Binary Decision Diagrams, to compactly represent and efficiently search the automaton. In this paper we make the following contributions. First, we present a general planning algorithm for conformant planning, which applies to fully nondeterministic domains, with uncertainty in the initial condition and in action effects. The algorithm is based on a breadth-first, backward search, and returns conformant plans of minimal length, if a solution to the planning problem exists, otherwise it terminates concluding that the problem admits no conformant solution. Second, we provide a symbolic representation of the search space based on Binary Decision Diagrams (BDDs), which is the basis for search techniques derived from symbolic model checking. The symbolic representation makes it possible to analyze potentially large sets of states and transitions in a single computation step, thus providing for an efficient implementation. Third, we present CMBP (Conformant Model Based Planner), an efficient implementation of the data structures and algorithm described above, directly based on BDD manipulations, which allows for a compact representation of the search layers and an efficient implementation of the search steps. Finally, we present an experimental comparison of our approach with the state-of-the-art conformant planners CGP, QBFPLAN and GPT. Our analysis includes all the planning problems from the distribution packages of these systems, plus other problems defined to stress a number of specific factors. Our approach appears to be the most effective: CMBP is strictly more expressive than QBFPLAN and CGP and, in all the problems where a comparison is possible, CMBP outperforms its competitors, sometimes by orders of magnitude.

## 1. Introduction

In recent years, there has been a growing interest in planning in nondeterministic domains. Rejecting some fundamental (and often unrealistic) assumptions of classical planning, domains are considered where actions can have uncertain effects, exogenous events are possible, and the initial state can be only partly specified. The challenge is to find a *strong plan*, that is guaranteed to achieve the goal despite the nondeterminism of the domain, regardless of the uncertainty on the initial condition and on the effect of actions. Conditional planning (Cassandra, Kaelbling, & Littman, 1994; Weld, Anderson, & Smith, 1998; Cimatti, Roveri, & Traverso, 1998b) tackles this problem by searching for a conditional course of actions, that depends on information that can be gathered at run-time. In certain domains, however, run-time information gathering may be too expensive or simply impossible. *Conformant planning* (Goldman & Boddy, 1996) is the problem of finding an unconditioned course of actions, i.e. a classical plan, that does not depend on run-time information gathering to guarantee the achievement of the goal. Conformant planning has been recognized as a significant problem in Artificial Intelligence since the work by Michie (1974): the Blind Robot problem requires to program the activity for a sensorless agent, which can be positioned in any location of a given room, so that it will be guaranteed to achieve a given goal. Conformant planning can be also seen as a problem of control for a system with an unobservable and unknown state, such as a microprocessor at power-up, or a software system under black-box testing.

Because of uncertainty, a plan is associated to potentially many different executions, which must be all taken into account in order to guarantee goal achievement. This makes conformant planning significantly harder than classical planning (Rintanen, 1999a; De Giacomo & Vardi, 1999). Despite this increased complexity, several approaches to conformant planning have been recently proposed, based on (extensions of) the main planning techniques for classical planning. The most interesting are CGP (Smith & Weld, 1998) based on GRAPHPLAN, QBFPLAN (Rintanen, 1999a) which extends the SAT-plan approach to QBF, and GPT (Bonet & Geffner, 2000) which encodes conformant planning as heuristic search. In this paper, we propose a new approach to conformant planning, based on Symbolic Model Checking (McMillan, 1993). Symbolic Model Checking is a formal verification technique, which allows one to analyze finite state automata of high complexity, relying on symbolic techniques, Binary Decision Diagrams (BDDs) (Bryant, 1986) in particular, for the compact representation and efficient search of the automaton. Our approach builds on the planning via model checking paradigm presented by Cimatti and his colleagues (1997, 1998b, 1998a), where finite state automata are used to represent complex, nondeterministic planning domains, and planning is based on (extensions of) the basic model checking steps. We make the following contributions.

- First, we present a general algorithm for conformant planning, which applies to any nondeterministic domain with uncertain action effects and initial condition, expressed as a nondeterministic finite-state automaton. The algorithm performs a breadth-first search, exploring plans of increasing length, until a plan is found or no more candidate plans are available. The algorithm is complete, i.e. it returns with failure if and only if the problem admits no conformant solution. If the problem admits a solution, the algorithm returns a conformant plan of minimal length.

- Second, we provide a symbolic representation of the search space based on Binary Decision Diagrams, which allows for the application of search techniques derived from symbolic model checking. The symbolic representation makes it possible to analyze *sets* of transitions in a single computation step. These sets can be compactly represented and efficiently manipulated despite their potentially large cardinality. This way it is possible to overcome the enumerative nature of the other approaches to conformant planning, for which the degree of nondeterminism tends to be a limiting factor.

- Third, we developed CMBP (Conformant Model Based Planner), which is an efficient implementation of the data structures and algorithm described above. CMBP is developed on top of MBP, the planner based on symbolic model checking techniques developed by Cimatti, Roveri and Traveso (1998b, 1998a). CMBP implements several new techniques, directly based on BDD manipulations, to compact the search layers and optimize termination checking.

- Finally, we provide an experimental evaluation of the state-of-the-art conformant planners, comparing CMBP with CGP, QBFPLAN and GPT. Because of the difference in expressivity, not all the problems which can be tackled by CMBP can also be represented in the other planners. However, for the problems where a direct comparison was possible, CMBP outperforms its competitors. In particular, it features a better qualitative behavior, not directly related to the *number* of initial states and uncertain action effects, and more stable with respect to the use of heuristics.

The paper is structured as follows. In Section 2 we review the representation of (non-deterministic) planning domains as finite state automata. In Section 3 we provide the intuitions and a formal definition of conformant planning in this setting. In Section 4 we present the planning algorithm, and in Section 5 we discuss the symbolic representation of the search space, which allows for an efficient implementation. In Section 6 we present the CMBP planner, and in Section 7 we present the experimental results. In Section 8 we discuss some further related work. In Section 9 we draw the conclusions and discuss future research directions.

## 2. Planning Domains as Finite State Automata

We are interested in complex, nondeterministic planning domains, where actions can have preconditions, conditional effects, and uncertain effects, and the initial state can be only partly specified. In the rest of this paper, we use a very simple though paradigmatic domain for explanatory purposes, a variation of Moore's *bomb in the toilet* domain (McDermott, 1987) (from now on called BTUC — BT with Uncertain Clogging). There are two packages, and one of them contains an armed bomb. It is possible to dunk either package in the toilet (actions $Dunk_1$ and $Dunk_2$), provided that the toilet is not clogged. Dunking either package has the uncertain effect of clogging the toilet. Furthermore, dunking the package containing the bomb has the effect of disarming the bomb. The action $Flush$ has the effect of unclogging the toilet.

We represent such domains as finite state automata. Figure 1 depicts the automaton for the BTUC domain. Each state is given a number, and contains all the propositions holding in that state. For instance, state 1 represents the state where the bomb is in package 1, is not defused, and the toilet is not clogged. Given that there is only one bomb, we write $In_2$ as an abbreviation for the negation of $In_1$. Arrows between states depict the transitions of the automaton, representing the possible behavior of actions. The transition from state 2 to state 1 labeled by $Flush$ represents the fact that the action $Flush$, if executed in state 2, only has the effect of removing the clogging. The execution of $Dunk_1$ in state 1, which has the uncertain effect of clogging the toilet, is represented by the multiple transitions to states 5 and 6. Since there is no transition outgoing from state 2 and labelled by $Dunk_1$, state 2 does not satisfy the preconditions of action $Dunk_1$, i.e. $Dunk_1$ is not applicable in state 2.

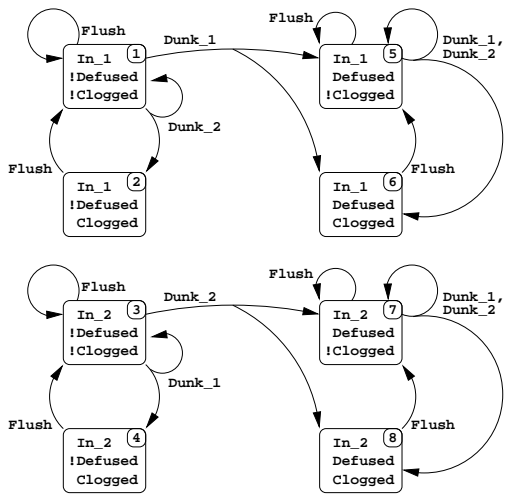

Figure 1: The automaton for the BTUC domain

We formally define nondeterministic planning domains as follows.

**Definition 1 (Planning Domain)** *A Planning Domain is a 4-tuple* $\mathcal{D} = (\mathcal{P}, \mathcal{S}, \mathcal{A}, \mathcal{R})$, *where* $\mathcal{P}$ *is the (finite) set of atomic propositions,* $\mathcal{S} \subseteq 2^{\mathcal{P}}$ *is the set of states,* $\mathcal{A}$ *is the (finite) set of actions, and* $\mathcal{R} \subseteq \mathcal{S} \times \mathcal{A} \times \mathcal{S}$ *is the transition relation.*

Intuitively, a proposition is in a state if and only if it holds in that state. In the following we assume that a planning domain $\mathcal{D}$ is given. We use $s$, $s'$ and $s''$ to denote states of $\mathcal{D}$, and $\alpha$ to denote actions. $\mathcal{R}(s, \alpha, s')$ holds iff when executing the action $\alpha$ in the state $s$ the state $s'$ is a possible outcome. We say that an action $\alpha$ is applicable in $s$ iff there is at least one state $s'$ such that $\mathcal{R}(s, \alpha, s')$ holds. We say that an action $\alpha$ is deterministic in $s$ iff there is a unique state $s'$ such that $\mathcal{R}(s, \alpha, s')$ holds. An action $\alpha$ has an uncertain outcome in $s$ if there are at least two distinct states $s'$ and $s''$ such that $\mathcal{R}(s, \alpha, s')$ and $\mathcal{R}(s, \alpha, s'')$ hold. As described by Cimatti and his colleagues (1997), the automaton for a given domain can be efficiently built starting from a compact description given in an expressive high level action language, for instance $\mathcal{AR}$ (Giunchiglia, Kartha, & Lifschitz, 1997).

## 3. Conformant Planning

Conformant planning (Goldman & Boddy, 1996) can be described as the problem of finding a sequence of actions that is guaranteed to achieve the goal regardless of the nondeterminism of the domain. That is, for *all* possible initial states, and for *all* uncertain action effects, the execution of the plan results in a goal state.

Consider the following problem for the BTUC domain. Initially, the bomb is armed but its position and the status of the toilet are uncertain, i.e. the initial state can be any of the states in $\{1, 2, 3, 4\}$. The goal is to reach a state where the bomb is defused, and the toilet

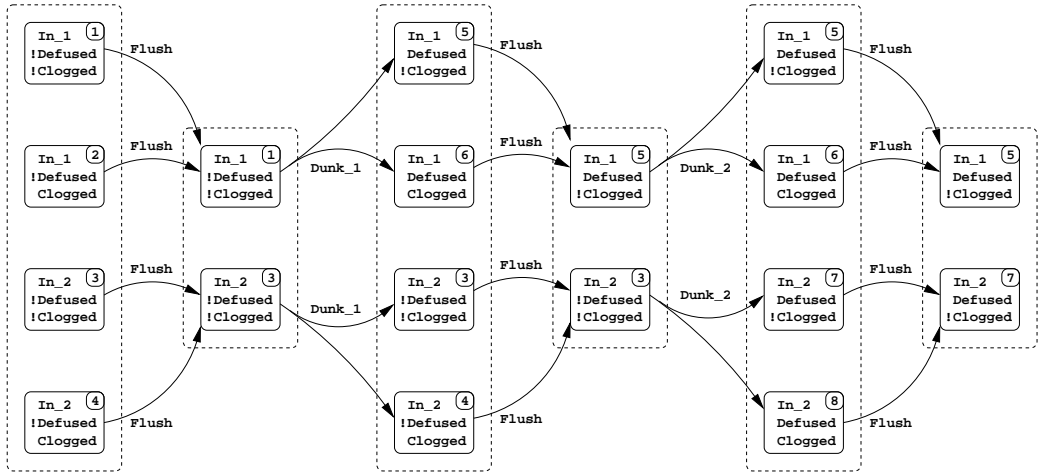

Figure 2: A conformant solution for the BTUC problem

is not clogged, i.e. the set of goal states is $\{5, 7\}$. A conformant plan solving this problem is

$$Flush;\ Dunk_1\ ;\ Flush\ ;\ Dunk_2;\ Flush \tag{1}$$

Figure 2 outlines the possible executions of the plan, for all possible initial states and uncertain action effects. The initial uncertainty lies in the fact that the domain might be in any of the states in $\{1, 2, 3, 4\}$. The possible initial states of the planning domain are collected into a set by a dashed line. We call such a set a *belief state*. Intuitively, a belief state expresses a condition of uncertainty about the domain, by collecting together all the states which are indistinguishable from the point of view of an agent reasoning about the domain. The first action, $Flush$, is used to remove the possible clogging. This reduces the uncertainty to the belief state $\{1, 3\}$. Despite the remaining uncertainty (i.e. it is still not known in which package the bomb is), action $Dunk_1$ is now guaranteed to be applicable because its precondition is met in both states. $Dunk_1$ has the effect of defusing the bomb if it is contained in package 1, and has the uncertain effect of clogging the toilet. The resulting belief state is $\{3, 4, 5, 6\}$. The following action, $Flush$, removes the clogging, reducing the uncertainty to the belief state $\{3, 5\}$, and guarantees the applicability of $Dunk_2$. After $Dunk_2$, the bomb is guaranteed to be defused, but the toilet might be clogged again (states 6 and 8 in the belief state $\{5, 6, 7, 8\}$). The final $Flush$ reduces the uncertainty to the belief state $\{5, 7\}$, and guarantees the achievement of the goal.

In general, in order for a plan to be a conformant solution, no action must be executed in states which do not satisfy the preconditions, and any state that can result from the execution of the plan (for all the initial states and for all the uncertain action effects) is a goal state. The main difficulty in achieving these conditions is that no information is (assumed to be) available at run-time. Therefore, at planning time we face the problem of reasoning about action execution in a belief state, i.e. under a condition of uncertainty.

**Definition 2 (Action Applicability)** *Let $Bs \subseteq \mathcal{S}$ be a Belief State. The action $\alpha$ is applicable in $Bs$ iff $Bs \neq \emptyset$ and $\alpha$ is applicable in every state $s \in Bs$.*

In order for an action to be applicable in a belief state, we require that its preconditions must be guaranteed notwithstanding the uncertainty. In other words, we reject "reckless" plans, which take the chance of applying an action without the guarantee of its applicability. This choice is strongly motivated in practical domains, where possibly fatal consequences can follow from the attempt to apply an action when its preconditions might not be satisfied (e.g. starting to fix an electrical device without being sure that it is not powered). The effect of action execution from an uncertain condition is defined as follows.

**Definition 3 (Action Image)** *Let $Bs \subseteq \mathcal{S}$ be a belief state, and let $\alpha$ be an action applicable in $Bs$. The image (also called execution) of $\alpha$ in $Bs$, written $Image[\alpha](Bs)$, is defined as follows.*

$$Image[\alpha](Bs) \quad \dot{=} \quad \{s' \mid \text{there exists } s \in Bs \text{ such that } \mathcal{R}(s, \alpha, s')\}$$

Notice that the image of an action combines the uncertainty in the belief state with the uncertainty on the action effects. (Consider for instance that $Image[Dunk_1](\{1,3\}) = \{3,4,5,6\}$.) In the following, we write $Image[\alpha](s)$ instead of $Image[\alpha](\{s\})$.

Plans are elements of $\mathcal{A}^*$, i.e. finite sequences of actions. We use $\epsilon$ for the 0-length plan, $\pi$ and $\rho$ to denote generic plans, and $\pi; \rho$ for plan concatenation. The notions of applicability and image generalize to plans as follows.

**Definition 4 (Plan Applicability and Image)** *Let $\pi \in \mathcal{A}^*$, and let $Bs \subseteq \mathcal{S}$. $\pi$ is applicable in $Bs$ iff one of the following holds:*

*1. $\pi = \epsilon$ and $Bs \neq \emptyset$;*

*2. $\pi = \alpha; \rho$, $\alpha$ is applicable in $Bs$, and $\rho$ is applicable in $Image[\alpha](Bs)$.*

*The image (also called execution) of $\pi$ in $Bs$, written $Image[\pi](Bs)$, is defined as:*

*1. $Image[\epsilon](Bs) \dot{=} Bs$;*

*2. $Image[\alpha; \pi](Bs) \dot{=} Image[\pi](Image[\alpha](Bs))$;*

A planning problem is formally characterized by the set of initial and goal states. The following definition captures the intuitive meaning of conformant plan given above.

**Definition 5 (Conformant Planning)** *Let $\mathcal{D} = (\mathcal{P}, \mathcal{S}, \mathcal{A}, \mathcal{R})$ be a planning domain. A Planning Problem for $\mathcal{D}$ is a triple $(\mathcal{D}, \mathcal{I}, \mathcal{G})$, where $\emptyset \neq \mathcal{I} \subseteq \mathcal{S}$ and $\emptyset \neq \mathcal{G} \subseteq \mathcal{S}$.*

*The plan $\pi$ is a conformant plan for (that is, a conformant solution to) the planning problem $(\mathcal{D}, \mathcal{I}, \mathcal{G})$ iff the following conditions hold:*

*(i) $\pi$ is applicable in $\mathcal{I}$;*

*(ii) $Image[\pi](\mathcal{I}) \subseteq \mathcal{G}$.*

In the following, when clear from the context, we omit the domain from the planning problem, and we simply write $(\mathcal{I}, \mathcal{G})$.

## 4. The Conformant Planning Algorithm

Our conformant planning algorithm is based on the exploration of the space of plans, limiting the exploration to plans which are conformant by construction. The algorithm builds Belief state-Plan (BsP) pairs of the form $\langle Bs \, . \, \pi \rangle$, where $Bs$ is a non-empty belief state and $\pi$ is a plan. The idea is to use a BsP pair to associate each explored plan with the maximal belief state where it is applicable, and from which it is guaranteed to result in goal states. The exploration is based on the basic function $SPreImage[\alpha](Bs)$, that, given a belief state $Bs$ and an action $\alpha$, returns the belief state containing all the states where $\alpha$ is applicable, and whose image under $\alpha$ is contained in $Bs$.

**Definition 6 (Strong Pre-Image)** *Let $\emptyset \neq Bs \subseteq \mathcal{S}$ be a belief state and let $\alpha$ be an action. The strong pre-image of $Bs$ under $\alpha$, written $SPreImage[\alpha](Bs)$, is defined as follows.*

$$SPreImage[\alpha](Bs) \doteq \{s \mid \alpha \text{ is applicable in } s, \text{and } Image[\alpha](s) \subseteq Bs\}$$

If $SPreImage[\alpha](Bs)$ is not empty, then $\alpha$ is applicable in it, and it is a conformant solution to the problem $(SPreImage[\alpha](Bs), Bs)$. Therefore, if the plan $\pi$ is a conformant solution for the problem $(Bs, \mathcal{G})$, then the plan $\alpha; \pi$ is a conformant solution to the problem $(SPreImage[\alpha](Bs), \mathcal{G})$.

Figure 3 depicts the space of BsP pairs built by the algorithm while solving the BTUC problem. The levels are built from the goal, on the right, towards the initial states, on the left. At level 0, the only BsP pair is $\langle \{5, 7\} \, . \, \epsilon \rangle$, composed by the set of goal states indexed by the 0-length plan $\epsilon$. (Notice that $\epsilon$ is a conformant solution to every problem with goal set $\{5, 7\}$ and initial states contained in $\{5, 7\}$.) The dashed arrows represent the application of $SPreImage$. At level 1, only the BsP pair $\langle \{5, 6, 7, 8\} \, . \, Flush \rangle$ is built, since the strong pre-image of the belief state 0 for the actions $Dunk_1$ and $Dunk_2$ is empty. At level 2, there are three BsP pairs, with (overlapping) belief states $Bs_2$, $Bs_3$ and $Bs_4$, indexed, respectively, by the length 2 plans $Dunk_1; Flush$, $Flush; Flush$ and $Dunk_2; Flush$. (A plan associated with a belief state $Bs_i$ is a sequence of actions labeling the path from $Bs_i$ to $Bs_0$.) Notice that $Bs_3$ is equal to $Bs_1$, and therefore deserves no further expansion. The expansion of belief states 2 and 4 gives the belief states 5 and 6, both obtained by the strong pre-image under $Flush$, while the strong pre-image under actions $Dunk_1$ and $Dunk_2$ returns empty belief states. The further expansion of $Bs_5$ results in three belief states. The one resulting from the strong pre-image under $Flush$ is not reported, since it is equal to $Bs_5$. Belief state 7 is also equal to $Bs_2$, and deserves no further expansion. Belief state 8 can be obtained by expanding both $Bs_5$ and $Bs_6$. At level 5, the expansion produces $Bs_{10}$, which contains all the initial states. Therefore, both of the corresponding plans are conformant solutions to the problem.

The conformant planning algorithm CONFORMANTPLAN is presented in Figure 4. It takes as input the planning problem in the form of the set of states $\mathcal{I}$ and $\mathcal{G}$ (the domain $\mathcal{D}$ is assumed to be globally available). The algorithm performs a backwards breadth-first search, exploring BsP pairs corresponding to plans of increasing length at each step. The status of the search (each level in Figure 3) is represented by a BsP table, i.e. a set of BsP pairs

$$BsPT = \{\langle Bs_1 \, . \, \pi_1 \rangle, \ldots, \langle Bs_n \, . \, \pi_n \rangle\}$$

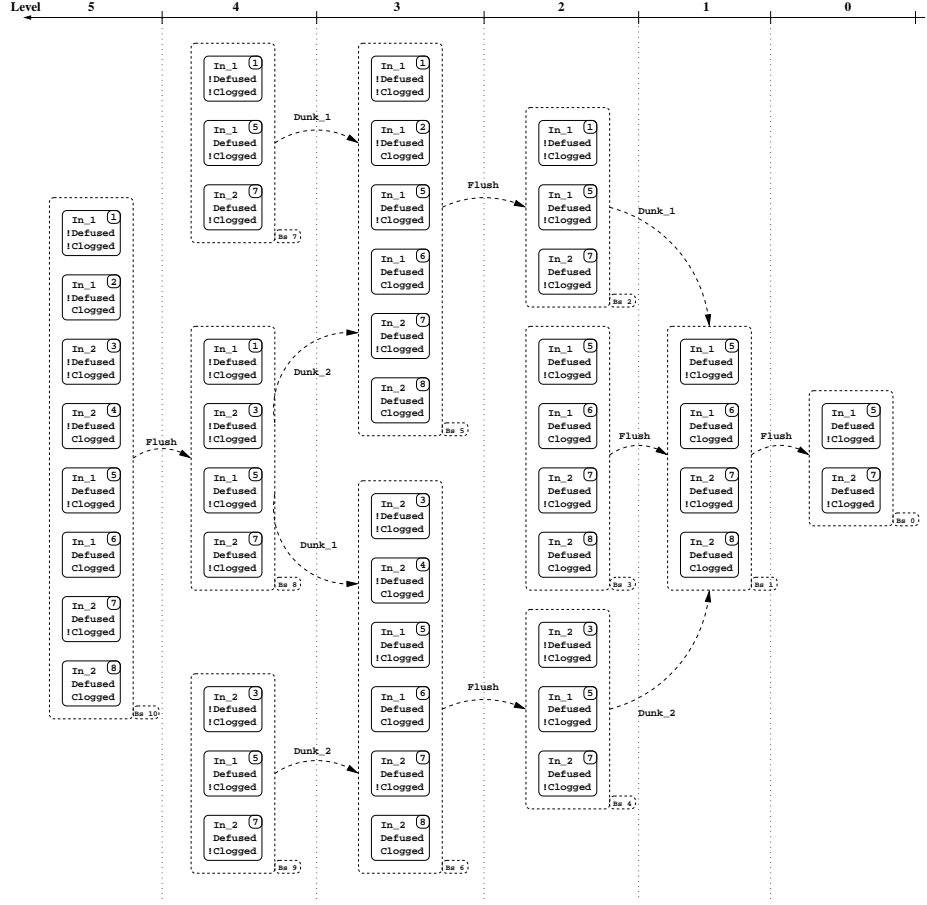

Figure 3: The BsP tables for the BTUC problem

where the $\pi_i$ are plans of the same length, such that $\pi_i \neq \pi_j$ for all $1 \leq j \neq i \leq n$. We call $Bs_i$ the belief set indexed by $\pi_i$. When no ambiguity arises, we write $BsPT(\pi_i)$ for $Bs_i$. The array $BsPTables$ is used to store the BsP tables representing the levels of the search. The algorithm first checks (line 4) if there are plans of length 0, i.e. if $\epsilon$ is a solution. If no conformant plan of such length exists (($Plans = \emptyset$) in line 4), then the while loop is entered. At each iteration, conformant plans of increasing length are explored (lines 5 to 8). The step at line 6 expands the BsP table in $BsPTables[i-1]$ and stores the resulting BsP table in $BsPTables[i]$. BsP pairs which are redundant with respect to the current search are eliminated from $BsPTables[i]$ (line 7). The possible solutions contained in $BsPTables[i]$ are extracted and stored in $Plans$ (line 8). The loop terminates if either a plan is found ($Plans \neq \emptyset$), or the space of conformant plans has been completely explored ($BsPTables[i] = \emptyset$).

The definitions of the basic functions used in the algorithm are reported in Figure 5. The function EXPANDBSPTABLE expands the BsP table provided as argument, containing conformant plans of length $i-1$, and returns a BsP table with conformant plans of length $i$. Each BsP in the input BsP table is expanded by EXPANDBSPPAIR. For each possible

```
     function ConformantPlan(I,G)
0    begin
1        i = 0;
2        BsPTables[0] := { ⟨G . ε⟩ };
3        Plans := ExtractSolution(I, BsPTables[0]);
4        while ((BsPTables[i] ≠ ∅) ∧ (Plans = ∅)) do
5            i := i + 1;
6            BsPTables[i] := ExpandBsPTable(BsPTables[i-1]);
7            BsPTables[i] := PruneBsPTable(BsPTables[i], BsPTables, i);
8            Plans := ExtractSolution(I, BsPTables[i]);
9        done
10       if (BsPTables[i] = ∅) then
11           return Fail;
12           else return Plans;
13   end
```

Figure 4: The conformant planning algorithm.

action $\alpha$, the strong pre-image of $Bs$ is computed, and if the resulting belief state $Bs'$ is not empty, i.e. there is a belief state from which $\alpha$ guarantees the achievement of $Bs$, then the plan $\pi$ is extended with $\alpha$ and $\langle Bs' \cdot \alpha; \pi \rangle$ is returned. The expansion of a BsP table is the union of the expansions of each BsP pair. The function ExtractSolution takes as input a BsP table and returns the (possibly empty) set of plans which index a belief states containing $I$. PruneBsPTable takes as input the BsP table to be pruned, an array of previously constructed BsP tables $BsPTables$, and an index of the current step. It removes from the BsP table in the input the plans which are not worth being explored because the corresponding belief states have already been visited.

The algorithm has the following properties. First, it always terminates. This follows from the fact that the set of explored belief sets (stored in $BsPTables$) is monotonically increasing — at each step we proceed only if at least one new belief state is generated. Because of its finiteness (the set of accumulated belief states is contained in $2^S$ which is finite), a fix point is eventually reached. Second, it is correct, i.e. when a plan is returned it is a conformant solution to the given problem. The correctness of the algorithm follows from the properties of $SPreImage$: each plan is associated with a belief state for which it is conformant, i.e. where it is guaranteed to be applicable and from which it results in a belief state contained in the goal. Third, the algorithm is optimal, i.e. it returns plans of minimal length. This property follows from the breadth-first style of the search. Finally, the algorithm is able to decide whether a problem admits no solution, returning $Fail$ in such cases. Indeed, a conformant solution is always associated with a belief state containing the initial states. $SPreImage$ generates the *maximal* belief state associated with a conformant plan, each new belief state generated in the exploration is compared with the initial states to check if it is a solution, and a plan is pruned only if an equivalent plan has already been explored.

$$\textsc{ExpandBsPTable}(BsPT) \doteq \bigcup_{\langle Bs \; . \; \pi \rangle \in BsPT} \textsc{ExpandBsPPair}(\langle Bs \; . \; \pi \rangle)$$

$$\textsc{ExpandBsPPair}(\langle Bs \; . \; \pi \rangle) \doteq \{\langle Bs' \; . \; \alpha ; \pi \rangle | \text{ such that } Bs' = SPreImage[\alpha](Bs) \neq \emptyset\}$$

$\textsc{PruneBsPTable}(BsPT, BsPTables, i) \doteq$
$\{\langle Bs \; . \; \pi \rangle \in BsPT \mid \text{ for all } j < i, \text{ there is no } \langle Bs \; . \; \pi' \rangle \in BsPTables[j] \text{ such that } (Bs' = Bs)\}$

$$\textsc{ExtractSolution}(\mathcal{I}, BsPT) \doteq \{\pi \mid \text{ there exists } \langle Bs \; . \; \pi \rangle \in BsPT \text{ such that } \mathcal{I} \subseteq Bs\}$$

Figure 5: The primitives used by the conformant planning algorithm.

# 5. Conformant Planning via Symbolic Model Checking

Model checking is a formal verification technique based on the exploration of finite state automata (Clarke, Emerson, & Sistla, 1986). Symbolic model checking (McMillan, 1993) is a particular form of model checking using Binary Decision Diagrams to compactly represent and efficiently analyze finite state automata. The introduction of symbolic techniques into model checking led to a breakthrough in the size of model which could be analyzed (Burch et al., 1992), and made it possible for model checking to be routinely applied in industry, especially in logic circuits design (for a survey see Clarke & Wing, 1996).

In the rest of this section, we will provide an overview of Binary Decision Diagrams, and we will describe the representation of planning domains, based on the BDD-based representation of finite state automata used in model checking. Then, we will discuss the extension which allows to symbolically represent BsP tables and their transformations, thus allowing for an efficient implementation of the algorithm described in the previous section.

## 5.1 Binary Decision Diagrams

A Reduced Ordered Binary Decision Diagram (Bryant, 1992, 1986) (improperly called BDD) is a directed acyclic graph (DAG). The terminal nodes are either $True$ or $False$. Each non-terminal node is associated with a boolean variable, and two BDDs, called left and right branches. Figure 6 (a) depicts a BDD for $(a_1 \leftrightarrow b_1) \wedge (a_2 \leftrightarrow b_2) \wedge (a_3 \leftrightarrow b_3)$. At each non-terminal node, the right [left, respectively] branch is depicted as a solid [dashed, resp.] line, and represents the assignment of the value $True$ [$False$, resp.] to the corresponding variable. A BDD represents a boolean function. For a given truth assignment to the variables in the BDD, the value of the function is determined by traversing the graph from the root to the leaves, following each branch indicated by the value assigned to the variables[1]. The reached leaf node is labeled with the resulting truth value. If $v$ is a BDD, its size $|v|$ is the number of its nodes. If $n$ is a node, $var(n)$ indicates the variable indexing node $n$.

---

1. A path from the root to a leaf can visit nodes associated with a subset of all the variables of the BDD. See for instance the path associated with $a_1, \neg b_1$ in Figure 6(a).

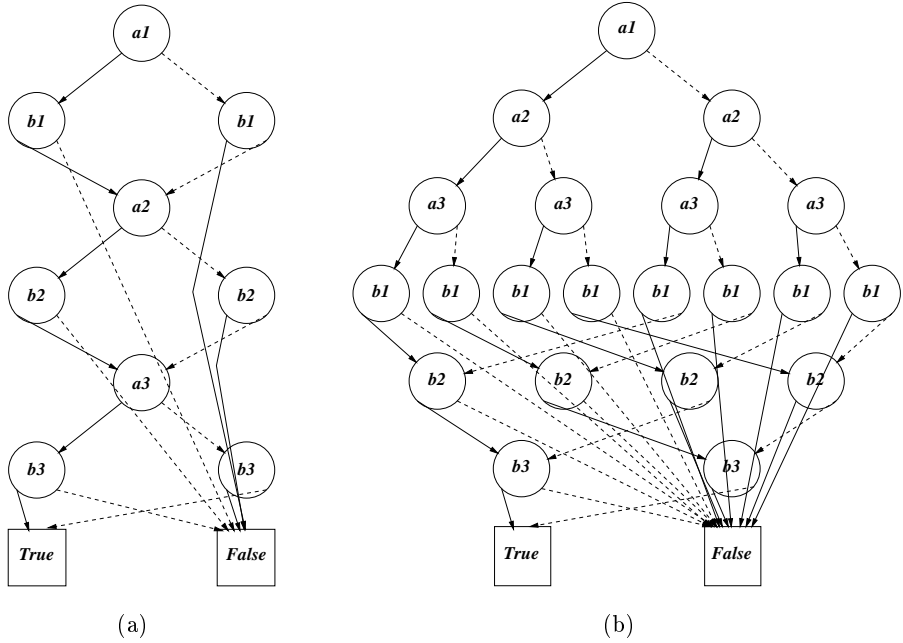

Figure 6: Two BDDs for the formula $(a_1 \leftrightarrow b_1) \land (a_2 \leftrightarrow b_2) \land (a_3 \leftrightarrow b_3)$.

BDDs are a canonical representation of Boolean functions. The canonicity follows by imposing a total order $<$ over the set of variables used to label nodes, such that for any node $n$ and respective non-terminal child $m$, their variables must be ordered, i.e. $var(n) < var(m)$, and requiring that the BDD contains no isomorphic subgraphs.

BDDs can be combined with the usual boolean transformations (e.g. negation, conjunction, disjunction). Given two BDDs, for instance, the conjunction operator builds and returns the BDD corresponding to the conjunction of its arguments. Substitution can also be represented as BDD transformations. In the following, if $v$ is a variable, and $\Phi$ and $\psi$ are BDDs, we indicate with $\Phi[v/\psi]$ the BDD resulting from the substitution of $v$ with $\psi$ in $\Phi$. If $\mathbf{v_1}$ and $\mathbf{v_2}$ are vectors of (the same number of) distinct variables, we indicate with $\Phi[\mathbf{v_1}/\mathbf{v_2}]$ the parallel substitution in $\Phi$ of the variables in vector $\mathbf{v_1}$ with the (corresponding) variables in $\mathbf{v_2}$.

BDDs also allow for transformations described as quantifications, in the style of Quantified Boolean Formulae (QBF). QBF is a definitional extension to propositional logic, where propositional variables can be universally and existentially quantified. In terms of BDD computations, a quantification corresponds to a tranformation mapping the BDD of $\Phi$ and the variable $v_i$ being quantified into the BDD of the resulting (propositional) formula. If $\Phi$ is a formula, and $v_i$ is one of its variables, the existential quantification of $v_i$ in $\Phi$, written $\exists v_i.\Phi(v_1,\ldots,v_n)$, is equivalent to $\Phi(v_1,\ldots,v_n)[v_i/False] \lor \Phi(v_1,\ldots,v_n)[v_i/True]$. Analogously, the universal quantification $\forall v_i.\Phi(v_1,\ldots,v_n)$ is equivalent to $\Phi(v_1,\ldots,v_n)[v_i/False] \land$ $\Phi(v_1, \ldots, v_n)[v_i/True]$. In QBF, quantifiers can be arbitrarily applied and nested. In general, a QBF formula has an equivalent propositional formula, but the conversion is subject to an exponential blow-up.

The time complexity of the algorithm for computing a truth-functional boolean transformation $f_1 <op> f_2$ is $O(|f_1| \cdot |f_2|)$. As far as quantifications are concerned, the time complexity is quadratic in the size of the BDD being quantified, and linear in the number of variables being quantified, i.e. $O(|\mathbf{v}| \cdot |f|^2)$ (Bryant, 1992, 1986).

BDD *packages* are efficient implementations of such data structures and algorithms (Brace et al., 1990; Somenzi, 1997; Yang et al., 1998; Coudert et al., 1993). Basically, a BDD package deals with a single multi-rooted DAG, where each node represents a boolean function. Memory efficiency is obtained by using a "unique table", and by sharing common subgraphs between BDDs. The unique table is used to guarantee that at each time there are no isomorphic subgraphs and no redundant nodes in the multi-rooted DAG. Before creating a new node, the unique table is checked to see if the node is already present, and only if this is not the case a new node is created and stored in the unique table. The unique table allows to perform the equivalence check between two BDDs in constant time (since two equivalent functions always share the same subgraph) (Brace et al., 1990; Somenzi, 1997). Time efficiency is obtained by maintaining a "computed table", which keeps track of the results of recently computed transformations, thus avoiding the recomputation.

A critical computational factor with BDDs is the order of the variables used. (Figure 6 shows an example of the impact of a change in the variable ordering on the size of a BDD.) For a certain class of boolean functions, the size of the corresponding BDD is exponential in the number of variables for any possible variable ordering (Bryant, 1991). In many practical cases, however, finding a good variable ordering is rather easy. Beside affecting the memory used to represent a Boolean function, finding a good variable ordering can have a big impact on computation times, since the complexity of the transformation algorithms depends on the size of the operands. Most BDD packages provide heuristic algorithms for finding good variable orderings, which can be called to try to reduce the overall size of the stored BDDs. The reordering algorithms can also be activated dynamically by the package, during a BDD computation, when the total number of nodes in the package reaches a predefined threshold (dynamic reoredering).

## 5.2 Symbolic Representation of Planning Domains

A planning domain $(\mathcal{P}, \mathcal{S}, \mathcal{A}, \mathcal{R})$ can be represented symbolically using BDDs, as follows. A set of (distinct) BDD variables, called *state* variables, is devoted to the representation of the states $\mathcal{S}$ of the domain. Each of these variables has a direct association with a proposition of the domain in $\mathcal{P}$ used in the description of the domain. For instance, for the BTUC domain, each of $In_1$, $Defused$ and $Clogged$ is associated with a unique BDD variable. In the following we write $\boldsymbol{x}$ for the vector of state variables. Because the particular order is irrelevant but for performance issues, in the rest of this section we will not distinguish a proposition and the corresponding BDD variable.

A state is a set of propositions of $\mathcal{P}$ (specifically, the propositions which are intended to hold in it). For each state $s$, there is a corresponding assignment to the state variables $\boldsymbol{x}$, i.e. the assignment where each variable corresponding to a proposition $p \in s$ is assigned to $True$, and each other variable is assigned to $False$. We represent $s$ with the BDD $\xi(s)$, having such an assignment as its unique satisfying assignment. For instance, $\xi(6) \doteq (In_1 \wedge Defused \wedge Clogged)$ is the BDD representing state 6, while $\xi(4) \doteq \neg In_1 \wedge \neg Defused \wedge Clogged$ represents state 4, and so on. (Without loss of generality, in the following we do not distinguish a propositional formula from the corresponding BDD.) This representation naturally extends to any $set\ of\ states\ Q \subseteq \mathcal{S}$ as follows:

$$\xi(Q) \doteq \bigvee_{s \in Q} \xi(s)$$

In other words, we associate a set of states with the generalized disjunction of the BDDs representing each of the states. Notice that the satisfying assignments of the $\xi(Q)$ are exactly the assignment representations of the states in $Q$. This representation mechanism is very natural. For instance, the BDD $\xi(\mathcal{I})$ representing the the set of initial states of the BTUC $\mathcal{I} \doteq \{1, 2, 3, 4\}$ is $\neg Defused$, while for the set of goal states $\mathcal{G} \doteq \{5, 7\}$ the corresponding BDD is $Defused \wedge \neg Clogged$. A BDD is also used to represent the set $\mathcal{S}$ of all the states of the domain automaton. In the BTUC, $\xi(\mathcal{S}) = True$ because $\mathcal{S} = 2^{\mathcal{P}}$. In a different formulation, where two $independent$ propositions $In_1$ and $In_2$ are used to represent the position of a bomb, $\xi(\mathcal{S})$ would be the BDD $In_1 \leftrightarrow \neg In_2$.

In general, a BDD represents the set of (states which correspond to) its models. As a consequence, set theoretic transformations are naturally represented by propositional operations, as follows.

$$
\begin{aligned}
\xi(\mathcal{S} \backslash Q) &\doteq \xi(\mathcal{S}) \wedge \neg\xi(Q) \\
\xi(Q_1 \cup Q_2) &\doteq \xi(Q_1) \vee \xi(Q_2) \\
\xi(Q_1 \cap Q_2) &\doteq \xi(Q_1) \wedge \xi(Q_2)
\end{aligned}
$$

The main efficiency of this symbolic representation lies in the fact that the cardinality of the represented set is not directly related to the size of the BDD. For instance, $\xi(\mathcal{G})$ uses two (non-terminal) nodes to represent two states, while $\xi(\mathcal{I})$ uses one node to represent four states. As limit cases, $\xi(\mathcal{S})$ and $\xi(\{\})$ are (the leaf BDDs) $True$ and $False$, respectively. As a further advantage, symbolic representation is extremely efficient in dealing with irrelevant information. Notice, for instance, that only the variable $Defused$ occurs in $\xi(\{5, 6, 7, 8\})$. For this reason, a symbolic representation can have a dramatic improvement over an explicit, enumerative representation. This is what allows symbolic, BDD-based model checkers to handle finite state automata with a very large number of states (see for instance Burch et al., 1992). In the following, we will collapse a set of states and the BDD representing it.

Another set of BDD variables, called $action$ variables, written $\boldsymbol{\alpha}$, is used to represent actions. We use one action variable for each possible action in $\mathcal{A}$. Intuitively, a BDD action variable is true if and only if the corresponding action is being executed. If we assume that a sequential encoding is used, i.e. no concurrent actions are allowed, we also use a BDD, $\text{SEQ}(\boldsymbol{\alpha})$, to express that exactly one of the action variables must be true at each time[2]. For the BTUC problem, where $\mathcal{A}$ contains three actions, we use the three BDD variables $Dunk_1$, $Dunk_2$ and $Flush$, while we express the serial encoding constraint with the following BDD:

$$\text{SEQ}(\boldsymbol{\alpha}) \doteq (Dunk_1 \vee Dunk_2 \vee Flush) \wedge \neg(Dunk_1 \wedge Dunk_2) \wedge \neg(Dunk_1 \wedge Flush) \wedge \neg(Dunk_2 \wedge Flush)$$

---

2. In the specific case of sequential encoding, an alternative approach using only $\lceil \log |\mathcal{A}| \rceil$ is possible: an assignment to the action variables denotes a specific action to be executed. Two assignments being mutually exclusive, the constraint $\text{SEQ}(\boldsymbol{\alpha})$ needs not to be represented. When the cardinality of the set of actions is not a power of two, the standard solution is to associate more than one assignment to certain values. This optimized solution, which is actually used in the implementation, is not described here for the sake of simplicity.

As for state variables, we are referring to BDD action variables with symbolic names for the sake of simplicity. In practice, they will be internally represented as integers, but their position in the ordering of the BDD package is totally irrelevant in logical terms.

A BDD in the variables $\boldsymbol{x}$ and $\boldsymbol{\alpha}$ represents a set of state-action pairs, i.e. a relation between states and actions. For instance, the applicability relation in the BTUC (i.e., all actions are possible in all states, except for dunking actions which require the toilet not to be clogged) is represented by the BDD $\neg(Clogged \wedge (Dunk_1 \vee Dunk_2))$. Notice that it represents a set of 16 state-action pairs, each associating a state with an applicable action.

A transition is a 3-tuple composed of a state (the initial state of the transition), an action (the action being executed), and a state (the resulting state of the transition). To represent transitions, another vector $\boldsymbol{x}'$ of BDD variables, called *next state* variables, is allocated in the BDD package. We write $\xi'(s)$ for the representation of the state $s$ in the next state variables. With $\xi'(Q)$ we denote the construction of the BDD corresponding to the set of states Q, using each variable in the next state vector $\boldsymbol{x}'$ instead of each current state variables $\boldsymbol{x}$. We require that $|\boldsymbol{x}| = |\boldsymbol{x}'|$, and assume that the $i$-th variable in $\boldsymbol{x}$ and the $i$-th variable in $\boldsymbol{x}'$ correspond. We define the representation of a set of states in the next variables as follows.

$$\xi'(s) \doteq \xi(s)[\boldsymbol{x}/\boldsymbol{x}']$$

We call the operation $\Phi[\boldsymbol{x}/\boldsymbol{x}']$ "forward shifting", because it transforms the representation of a set of "current" states in the representation of a set of "next" states. The dual operation $\Phi[\boldsymbol{x}'/\boldsymbol{x}]$ is called backward shifting. In the following, we call $\boldsymbol{x}$ *current* state variables to distinguish them from next state variables. A transition is represented as an assignment to $\boldsymbol{x}$, $\boldsymbol{\alpha}$ and $\boldsymbol{x}'$. For the BTUC, the transition corresponding to the application of action $Dunk_1$ in state 1 resulting in state 5 is represented by the following BDD

$$\xi(\langle 1, Dunk_1, 5 \rangle) \doteq \xi(1) \wedge Dunk_1 \wedge \xi'(5)$$

The transition relation $\mathcal{R}$ of the automaton corresponding to a planning domain is simply a set of transitions, and is thus represented by a BDD in the BDD variables $\boldsymbol{x}$, $\boldsymbol{\alpha}$ and $\boldsymbol{x}'$, where each satisfying assignment represents a possible transition.

$$\xi(\mathcal{R}) \doteq \text{SEQ}(\boldsymbol{\alpha}) \wedge \bigvee_{t \in \mathcal{R}} \xi(t)$$

In the rest of this paper, we assume that the BDD representation of a planning domain is given. In particular, we assume as given the vectors of variables $\boldsymbol{x}, \boldsymbol{x}', \boldsymbol{\alpha}$, the encoding functions $\xi$ and $\xi'$, and we simply call $\mathcal{S}$, $\mathcal{R}$, $\mathcal{I}$ and $\mathcal{G}$ the BDD representing the states of the domain, the transition relation, the initial states and the goal states, respectively. We write $\Phi(\mathbf{v})$ to stress that the BDD $\Phi$ depends on the variables in $\mathbf{v}$. With this representation, it is possible to reason about plans, simulating symbolically the execution of sets of actions in sets of states, by means of QBF transformations. The BDD representing the applicability relation can be directly obtained with the following computation.

$$\text{APPLICABLE}(\boldsymbol{x}, \boldsymbol{\alpha}) \doteq \exists \boldsymbol{x}'. \mathcal{R}(\boldsymbol{x}, \boldsymbol{\alpha}, \boldsymbol{x}')$$

The resulting Bdd, Applicable$(\boldsymbol{x}, \boldsymbol{\alpha})$, represents the set of state-action pairs such that the action is applicable in the state. The Bdd representing the states reachable from $Q$ in one step is obtained with the following computation.

$$\exists \boldsymbol{x}. \exists \boldsymbol{\alpha}. (\mathcal{R}(\boldsymbol{x}, \boldsymbol{\alpha}, \boldsymbol{x}') \wedge Q(\boldsymbol{x}))[\boldsymbol{x}'/\boldsymbol{x}]$$

Notice that, with this single operation, we symbolically simulate the effect of the application of any applicable action in $\mathcal{A}$ to any of the states in $Q$. Similarly, the following transformation allows to symbolically compute the *SPreImage* of a set of states $Q$ under all possible actions in one single computation:

$$\forall \boldsymbol{x}'. (\mathcal{R}(\boldsymbol{x}, \boldsymbol{\alpha}, \boldsymbol{x}') \rightarrow Q(\boldsymbol{x})[\boldsymbol{x}/\boldsymbol{x}']) \ \wedge \ \text{Applicable}(\boldsymbol{x}, \boldsymbol{\alpha})$$

The resulting Bdd represents all the state-action pairs $\langle \boldsymbol{x} \ . \ \boldsymbol{\alpha} \rangle$ such that $\boldsymbol{\alpha}$ is applicable in $\boldsymbol{x}$ and the execution of $\boldsymbol{\alpha}$ in $\boldsymbol{x}$ results in states in $Q$.

### 5.3 Symbolic Search in the Space of Belief States

The main strength of the symbolic approach is that it allows to perform a symbolic breadth-first search, and it provides a way for compactly representing and efficiently expanding the frontier. For instance, plans can be constructed by symbolic breadth-first search in the space of states, repeatedly applying the strong pre-image to the goal states (Cimatti et al., 1998b). However, the machinery presented in the previous section cannot be directly applied to tackle conformant planning. The basic difference is that with conformant planning we are searching in the space of belief states[3], and therefore the frontier of the search is basically a *set of sets of states*. We introduce a way to symbolically represent BsP tables. Basically, this can be seen as a construction on demand, based on the algorithm steps, of increasingly large portions of the space of belief states. The key intuition is that a BsP table

$$\{\langle \{s_1^1, \dots, s_{n_1}^1\} \ . \ \pi_1 \rangle, \dots, \langle \{s_1^k, \dots, s_{n_k}^k\} \ . \ \pi_k \rangle\}$$

is represented as a relation between plans (of the same length) and states, by associating the plan directly with each state in the belief state indexed by the plan, as follows:

$$\{\langle s_1^1 \ . \ \pi_1 \rangle, \dots, \langle s_{n_1}^1 \ . \ \pi_1 \rangle, \dots, \langle s_1^k \ . \ \pi_k \rangle, \dots, \langle s_{n_k}^k \ . \ \pi_k \rangle\} \tag{2}$$

We use additional variables to represent the plans in the BsP tables. In order to represent plans of increasing length, at each step of the algorithm, a vector of new Bdd variables, called *plan variables*, is introduced. The vector of plan variables introduced at the $i$-th step of the algorithm is written $\boldsymbol{\pi}_{[\mathbf{i}]}$, with $|\boldsymbol{\pi}_{[\mathbf{i}]}| = |\boldsymbol{\alpha}|$, and is used to encode the $i$-th to last action in the plan[4]. At step one of the algorithm, we introduce the vector of plan variables $\boldsymbol{\pi}_{[\mathbf{1}]}$ to represent the action corresponding to each 1-length possible conformant plan. The BsP

---

3. In principle, the machinery for symbolic search could be used to do conformant planning if applied to the determinization of the domain automaton, i.e. an automaton having $2^{\mathcal{S}}$ as its state space. However, this would require the introduction of an exponential number of state variables, which is impractical even for very small domains.

4. The search being performed backwards, plans need to be reversed once found.

table $BsPT_1$ at level 1 is built by EXPANDBSPTABLE by performing the following BDD computation starting from the BsP table at level 0, i.e. $\mathcal{G}(\boldsymbol{x})$:

$$(\forall \boldsymbol{x}'.(\mathcal{R}(\boldsymbol{x}, \boldsymbol{\alpha}, \boldsymbol{x}') \rightarrow \mathcal{G}(\boldsymbol{x})[\boldsymbol{x}/\boldsymbol{x}']) \;\wedge\; \text{APPLICABLE}(\boldsymbol{x}, \boldsymbol{\alpha}))[\boldsymbol{\alpha}/\boldsymbol{\pi}_{[1]}]$$

The computation collects those state-action pairs $\langle \boldsymbol{x} \,.\, \boldsymbol{\alpha} \rangle$ such that (the action represented by) $\boldsymbol{\alpha}$ is applicable in (the state represented by) $\boldsymbol{x}$, and such that all the resulting (states represented by) $\boldsymbol{x}'$ are goal states. Then we replace the vector of action variables $\boldsymbol{\alpha}$ with the first vector of plan variables $\boldsymbol{\pi}_{[1]}$. The resulting BDD, $BsPT(\boldsymbol{x}, \boldsymbol{\pi}_{[1]})$, represents a BsP table containing plans of length one in the form of a relation between states and plans as in (2). In the general case, after step $i - 1$, the BsP table $BsPT_{i-1}$, associating belief states to plans of length $i - 1$, is represented by a BDD in the state variables $\boldsymbol{x}$ and in the plan variables $\boldsymbol{\pi}_{[i-1]}, \ldots, \boldsymbol{\pi}_{[1]}$. The computation performed by EXPANDBSPTABLE at step $i$ is implemented as the following BDD transformation on $BsPT_{i-1}$

$$(\forall \boldsymbol{x}'.(\mathcal{R}(\boldsymbol{x}, \boldsymbol{\alpha}, \boldsymbol{x}') \rightarrow BsPT_{i-1}(\boldsymbol{x}, \boldsymbol{\pi}_{[i-1]}, \ldots, \boldsymbol{\pi}_{[1]})[\boldsymbol{x}/\boldsymbol{x}']) \;\wedge\; \text{APPLICABLE}(\boldsymbol{x}, \boldsymbol{\alpha}))[\boldsymbol{\alpha}/\boldsymbol{\pi}_{[i]}] \quad (3)$$

The next state variables in $\mathcal{R}$ and in $BsPT_{i-1}$ (resulting from the forward shifting) disappear because of the universal quantification. The action variables $\boldsymbol{\alpha}$ are renamed to the newly introduced plan variables $\boldsymbol{\pi}_{[i]}$, so that in the next step of the algorithm the construction can be repeated.

EXTRACTSOLUTION extracts the assignments to plan variables such that the corresponding set contains the initial states. In terms of BDD transformations, EXTRACTSOLUTION is implemented as follows:

$$\forall \boldsymbol{x}.(\mathcal{I}(\boldsymbol{x}) \rightarrow BsPT_i(\boldsymbol{x}, \boldsymbol{\pi}_{[i]}, \ldots, \boldsymbol{\pi}_{[1]})) \quad (4)$$

The result is a BDD in the plan variables $\boldsymbol{\pi}_{[i]}, \ldots, \boldsymbol{\pi}_{[1]}$. If the BDD is $False$, then there are no solutions of length $i$. Otherwise, each of the satisfying assignments to the resulting BDD represents a conformant solution to the problem.

To guarantee the termination of the algorithm, at each step the BsP table returned by EXPANDBSPTABLE is simplified by PRUNEBSPTABLE by removing all the belief states which do not deserve further expansion. This requires the comparison of the belief states contained in the BsP table with the belief states contained in each of the BsP tables built at previous levels. This is one of the crucial steps in terms of efficiency. An earlier implementation of this step with logical BDD transformations, following directly from the set-theoretical definition of PRUNEBSPTABLE, was extremely inefficient (Cimatti & Roveri, 1999). Furthermore, we noticed that the serial encoding could yield BsP tables containing a large number of equivalent plans, all indexing exactly the same belief state. Often these equivalent plans only differ in the order of some independent actions, and this is a potential source of combinatorial explosion. This occurs even in the simple version of the BTUC (in Figure 3, two equivalent conformant plans are associated with $Bs_8$). Therefore, we developed a new implementation which could tackle these two problems by operating directly on the BsP table. The idea is depicted in Figure 7. Initially, the cache contains $Bs_1$, $Bs_2$ and $Bs_3$. The simplification performs a traversal of the BDD, by accumulating the subtrees representing belief states, comparing them with the ones built at previous levels, and inserting the new ones in the cache (in Figure 7, $Bs_4$, $Bs_5$ and $Bs_6$). Each time a path is identified which represents a plan indexing an already cached belief state, the plan is redundant and the corresponding path is pruned[5]. The cost of the simplification is linear in the size of the BsP being simplified and is highly effective in pruning.

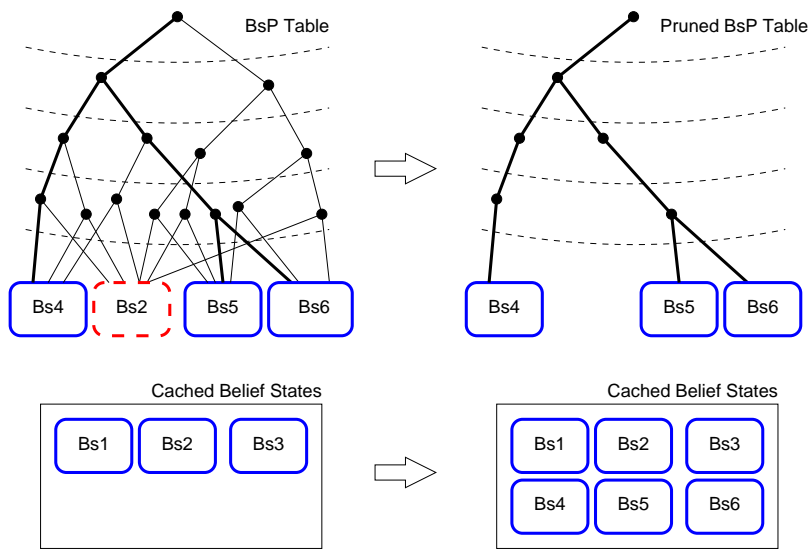

Figure 7: An example of pruning of a BsP table

# 6. CMBP: a BDD-based Conformant Planner

CMBP (Conformant Model Based Planner) is a conformant planner implementing the data structures and algorithms for conformant planning described in the previous sections. CMBP inherits the features of MBP (Cimatti et al., 1997, 1998b, 1998a), a planner based on symbolic model checking techniques. MBP is built on top of NUSMV, a symbolic model checker jointly developed by ITC-IRST and CMU (Cimatti et al., 2000), and uses the CUDD (Somenzi, 1997) state-of-the-art BDD package. MBP is a two-stage system. In the first stage, an internal BDD-based representation of the domain is built, while in the second stage planning problems can be solved. Currently, planning domains are described by means of the high-level action language $\mathcal{AR}$ (Giunchiglia et al., 1997). $\mathcal{AR}$ allows to specify (conditional and uncertain) effects of actions by means of high level assertions. For instance, Figure 8 shows the $\mathcal{AR}$ description of the BTUC problem[6]. The semantics of $\mathcal{AR}$ yields a serial encoding, i.e. exactly one action is assumed to be executed at each

---

5. This pruning mechanism is actually weaker than the earlier one (Cimatti & Roveri, 1999). Here we require that the same belief state must not be expanded twice during the search, while in the earlier version we prune belief states *contained* in previously explored ones. This may increase the *number* of explored belief states. However, it allows for a much more efficient implementation, without impacting on the properties of the algorithm.

6. ! and & stand for negation and conjunction, respectively. The description is slightly edited for the sake of readability. In particular, MBP currently does not accept parameterized $\mathcal{AR}$ descriptions. In practice we use a script language to generate ground instances of different complexity from a parameterized problem description.

```
DOMAIN BTUC

ACTIONS Dunk_1, Dunk_2, Flush;
FLUENTS In_1, In_2, Defused, Clogged : boolean;
INERTIAL Clogged, Defused, In_1, In_2;

ALWAYS In_1 <-> !In_2;

Flush CAUSES !Clogged;

for i in [1, 2] {
  Dunk_<i> HAS PRECONDITIONS !Clogged;
  Dunk_<i> CAUSES Defused IF In_<i>;
  Dunk_<i> POSSIBLY CHANGES Clogged;
}

INITIALLY !Defused;
CONFORMANT Defused & !Clogged;
```

Figure 8: An $\mathcal{AR}$ description for the BTUC problem

time. The automaton corresponding to an $\mathcal{AR}$ description is obtained by means of the minimization procedure by Giunchiglia (1996). This procedure solves the frame problem and the ramification problem, and is efficiently implemented in MBP (Cimatti et al., 1997). Because of the separation between the domain construction and the planning phases, MBP is not bound to $\mathcal{AR}$. Standard deterministic domains specified in PDDL (Ghallab et al., 1998) can also be given to MBP by means of a (prototype) compiler. We are also starting to investigate the potential use of the $\mathcal{C}$ action language (Giunchiglia & Lifschitz, 1998), which allows to represent domains with parallel actions.

Different planning algorithms can be applied to the specified planning problems. They operate solely on the automaton representation, and are completely independent of the particular language used to specify the domain. MBP allows for automatic construction of conditional plans under total observability, by implementing the algorithms for strong planning (Cimatti et al., 1998b), and for strong cyclic plannig (Cimatti et al., 1998a; Daniele, Traverso, & Vardi, 1999). In CMBP, we implemented the ideas described in the previous sections. The primitives to construct and prune BsP tables required a lot of tuning, in particular with the ordering of BDD variables. We found a general ordering strategy which works reasonably well: action variables are positioned at the top of the ordering, followed by plan variables, followed by state variables, with current state and next state variables interleaved. The specific ordering within action variables, plan variables, and state variables is determined by the standard mechanism implemented in NuSMV. CMBP implements several algorithms for conformant planning. In addition to the backward algorithm presented in Section 4, CMBP implements an algorithm based on forward search, which allows to exploit the initial knowledge of the problem, sometimes resulting in significant speed ups (Cimatti & Roveri, 2000). Backward and forward search can also be combined, to tackle the exponential growth of the search time with the depth of search. For all these algorithms, different options enable and disable different versions of the termination check.

## 7. Experimental Evaluation

In this section we present an experimental evaluation of our approach, which was carried out by comparing CMBP with state-of-the-art conformant planners. We first describe the other conformant planners considered in the analysis, and then we present the experimental comparison that was carried out.

### 7.1 Other Conformant Planners

CGP (Smith & Weld, 1998) extends the ideas of GRAPHPLAN (Blum & Furst, 1995, 1997) to deal with uncertainty. Basically, a planning graph is built of every possible sequence of possible worlds, and constraints among planning graphs are propagated to ensure conformance. The CGP system takes as input domains described in an extension of PDDL (Ghallab et al., 1998), where it is possible to specify uncertainty in the initial state. CGP inherits from GRAPHPLAN the ability to deal with parallel actions. CGP was the first efficient conformant planner: it was shown to outperform several other planners such as Buridan (Peot, 1998) and UDTPOP (Kushmerick, Hanks, & Weld, 1995). The detailed comparison reported by Smith and Weld (1998) leaves no doubt on the superiority of CGP with respect to these systems. Therefore, we compared CMBP with CGP and did not consider the other systems analyzed by Smith and Weld (1998). CMBP is more expressive than CGP in two respects. First, CGP can only handle uncertainty in the initial state. For instance, CGP cannot analyze the BTUC domain presented in Section 3. Smith and Weld (1998) describe how the approach can be extended to actions with uncertain effects. Second, CGP cannot conclude that a planning problem has no conformant solutions.

QBFPLAN is (our name for) the planning system by Rintanen (1999a). QBFPLAN generalizes the idea of SAT-based planning (Kautz, McAllester, & Selman, 1996; Kautz & Selman, 1996, 1998) to nondeterministic domains, by encoding problems in QBF. The QBFPLAN approach is not limited to conformant planning, but can be used to do conditional planning under uncertainty, also under partial observability: different encodings, corresponding to different structures in the resulting plan, can be synthesized. In this paper, we are only considering encodings which enforce the resulting plan to be a sequence. Given a bound on the length of the plan, first a QBF encoding of the problem is generated, and then a QBF solver (Rintanen, 1999b) is called. If no solution is found, a new encoding for a longer plan must be generated and solved. QBFPLAN is able to handle actions with uncertain effects. This is done by introducing auxiliary (choice) variables, the assignments to which the different possible outcomes of actions correspond. These variables are universally quantified to ensure conformance of the solution. Differently from e.g. BLACKBOX (Kautz & Selman, 1998), QBFPLAN does not have a heuristic to guess the "right" length of the plan. Given a limit in the length of the plan, it generates all the encodings up to the specified length, and repeatedly calls the QBF solver on encodings of increasing length until a plan is found.

As CGP, QBFPLAN cannot conclude that a planning problem has no conformant solutions. Similarly to CMBP, QBFPLAN relies on a symbolic representation of the problem, although QBF transformations are performed by a theorem prover rather than with BDDs.

GPT (Bonet & Geffner, 2000) is a general planning framework, where the conformant planning problem is seen as deterministic search problem in the space of belief states. GPT uses an explicit representation of the search space, where each belief state is represented as a separate data structure. The search is based on the A* algorithm (Nilsson, 1980), driven by domain dependent heuristics which are automatically generated from the problem description. GPT accepts problem descriptions in a syntax based on PDDL, extended to deal with probabilities and uncertainty. It is possible to represent domains with uncertain action effects (although the representation of actions resulting in a large number of different states is rather awkward). As for the planning algorithm, GPT is able to conclude that a given planning problem has no conformant solution by exhaustively exploring the space of belief states.

## 7.2 Experiments and Results

The evaluation was performed by running the systems on a number of parameterized problem domains. We considered all the problems from the CGP and GPT distributions, plus other problems which were defined to test specific features of the planners. We considered domains with uncertainty limited to the initial state, and domains with uncertain action effects. Besides problems admitting a solution, we also considered problems not admitting a solution, in which case we measured the effectiveness of the plannner in returning with failure.

Given their different expressivity, it was not possible to run all the systems on all the examples. CMBP was run on all the classes of examples, while GPT was run on all but one. CGP was run only on the problems which admit a solution, and with uncertainty limited to the initial condition. QBFPLAN was run on all the examples for which an encoding was already available from the QBFPLAN distribution. This is only a subset of the problems expressible in CGP. The main limiting factor was the low level of the input format of QBFPLAN: problem descriptions must be specified as ML code which generates the QBF encodings. Writing new encodings turned out to be a very difficult task, especially due to the lack of documentation.

We ran CGP, QBFPLAN and CMBP on an Intel 300MHz Pentium-II, 512MB RAM, running Linux. The comparison between CMBP and GPT was run on a Sun Ultra Sparc 270MHz, 128Mb RAM running Solaris (GPT was available as a binary). However, the performance of the two machines is comparable — the run times for CMBP were almost identical. CPU time was limited to 7200 sec (two hours) for each test. To avoid swapping, the memory limit was fixed to the physical memory of the machine. In the following, we write "—" or "=" for a test that did not complete within the above time and memory limits, respectively. The performance of the systems are reported in tables listing only the search time. This excludes the time needed by QBFPLAN to generate the encodings, the time spent by CMBP to construct the automaton representation into BDD, and the time needed by GPT to generate the source code of its internal representation, and to compile it. Overall, the most significant time ignored is the automaton construction of CMBP.

Currently, the automaton construction is not fully optimized. Even in the most complex examples, however, the construction never required more than a couple of minutes[7].

### 7.2.1 Bomb in the Toilet

**Bomb in the Toilet.** The first domain we tackled is the classical bomb in the toilet, where there is no notion of clogging. We call the problem BT($p$), where the parameter $p$ is the number of packages. The only uncertainty is in the initial condition, where it is not known which package contains the bomb. The goal is to defuse the bomb. The results for the BT problem are shown in Table 1. The columns relative to Cmbp are the length of the plan ($|P|$), the number of cached belief states and the number of hits in the cache (#BS and #NBS respectively), the time (expressed in seconds) needed for searching the automaton under Pentium/Linux (Time(L)) and under Sparc/Solaris (Time(S)). In the following, when clear from the context, the execution platform is omitted. The columns relative to Cgp are the number of levels in the planning graphs ($|L|$) and the search time. The column relative to Gpt is the search time.

| | | Cmbp | | | | Cgp | | Gpt |
|---|---|---|---|---|---|---|---|---|
| | $|P|$ | #BS/#BSH | Time(L) | Time(S) | | $|L|$ | Time | Time |
| BT(2) | 2 | 2 / 2 | 0.000 | 0.000 | | 1 | 0.000 | 0.074 |
| BT(3) | 3 | 6 / 11 | 0.000 | 0.000 | | 1 | 0.000 | 0.077 |
| BT(4) | 4 | 14 / 36 | 0.000 | 0.000 | | 1 | 0.000 | 0.080 |
| BT(5) | 5 | 30 / 103 | 0.000 | 0.000 | | 1 | 0.000 | 0.087 |
| BT(6) | 6 | 62 / 266 | 0.010 | 0.010 | | 1 | 0.010 | 0.102 |
| BT(7) | 7 | 126 / 641 | 0.010 | 0.030 | | 1 | 0.010 | 0.139 |
| BT(8) | 8 | 254 / 1496 | 0.030 | 0.030 | | 1 | 0.020 | 0.230 |
| BT(9) | 9 | 510 / 3463 | 0.070 | 0.070 | | 1 | 0.020 | 0.481 |
| BT(10) | 10 | 1022 / 7862 | 0.150 | 0.140 | | 1 | 0.020 | 1.018 |

Table 1: Results for the BT problems.

The BT problem is intrinsically parallel, i.e. the depth of the planning graph is always one, because all the packages can be dunked at the same time. Cgp inherits from Graph-plan the ability to deal with parallel actions efficiently, and therefore it is almost insensitive to the problem size. For this problem Cgp outperforms both Cmbp and Gpt. Notice that the number of levels explored by Cgp is always 1, while the length of the plan produced by Cmbp and Cgp grows linearly. Cmbp performs slightly better than Gpt.

**Bomb in the Toilet with Clogging.** We call BTC($p$) the extension of the BT($p$) where dunking a package (always) clogs the toilet, flushing can remove the clogging, and no clogging is a precondition for dunking a package. Again, $p$ is the number of packages. The toilet is initially not clogged. With this modification, the problem no longer allows for a parallel solution. The results for this problem are listed in Table 2. The impact of the depth of the plan length becomes significant for all systems. Both Cmbp and Gpt outperform Cgp. In this case Cmbp performs better than Gpt, especially on large instances (see BTC(16)).

---

7. More precisely, the maximum time in building the automaton was required for the BMTC(10,6) examples (88 secs.), the RING(10) example (77 secs.), the BMTC(9,6) examples (40 secs.), and the BMTC(10,5) examples (41 secs.). For most of the other examples, the time required for the automaton construction was less than 10 seconds.

| | Qbfplan | | |
|---|---|---|---|
| | BTC(6) | | BTC(10) |
| |P | Time | |P | Time |
| 1 | 0.00 | 1 | 0.02 |
| 2 | 0.01 | 2 | 0.03 |
| 3 | 0.26 | 3 | 0.78 |
| 4 | 0.63 | 4 | 2.30 |
| 5 | 1.53 | 5 | 4.87 |
| 6 | 2.82 | 6 | 8.90 |
| 7 | 6.80 | 7 | 22.61 |
| 8 | 14.06 | 8 | 52.72 |
| 9 | 35.59 | 9 | 156.12 |
| 10 | 93.34 | 10 | 410.86 |
| 11 | (+) 2.48 | 11 | 1280.88 |
| | | 13 | 3924.96 |
| | | 14 | — |
| | | ... | ... |
| | | 18 | — |
| | | 19 | (+) 16.84 |

| | | Cmbp | | | | Cgp | | Gpt |
|---|---|---|---|---|---|---|---|---|
| | |P | #BS/#BSH | Time(L) | Time(S) | |L | Time | Time |
| BTC(2) | 3 | 6 / 8 | 0.000 | 0.010 | 3 | 0.000 | 0.074 |
| BTC(3) | 5 | 14 / 23 | 0.000 | 0.000 | 5 | 0.010 | 0.077 |
| BTC(4) | 7 | 30 / 61 | 0.010 | 0.010 | 7 | 0.030 | 0.082 |
| BTC(5) | 9 | 62 / 150 | 0.020 | 0.020 | 9 | 0.130 | 0.094 |
| BTC(6) | 11 | 126 / 347 | 0.020 | 0.020 | 11 | 0.860 | 0.113 |
| BTC(7) | 13 | 254 / 796 | 0.070 | 0.080 | 13 | 2.980 | 0.166 |
| BTC(8) | 15 | 510 / 1844 | 0.150 | 0.160 | 15 | 13.690 | 0.288 |
| BTC(9) | 17 | 1022 / 4149 | 0.320 | 0.330 | 17 | 41.010 | 0.607 |
| BTC(10) | 19 | 2046 / 9190 | 0.710 | 0.700 | 19 | 157.590 | 1.309 |
| BTC(16) | 31 | 131070 / 921355 | 99.200 | 99.800 | | | 351.457 |

Table 2: Results for the BTC problems.

The comparison with Qbfplan is limited to the 6 and 10 package instances (the ones available from the distribution package). The performance of Qbfplan is reported in the left table in Table 2. Each line reports the time needed to decide whether there is a plan of length $i$. The performance of Qbfplan is rather good when tackling an encoding admitting a solution (in Table 2 these entries are labeled by $(+)$). For instance, in the BTC(10) Qbfplan finds the solution solving the encodings at depth 19 reasonably fast. However, when a solution cannot be found, i.e. the QBF formula admits no model, the performance of Qbfplan degrades significantly (for the depth 18 encoding, we let the solver run for 10 CPU hours and it did not complete the search). Because of the difference in performance, and the difficulty in writing new domains, in the rest of the comparison we will not consider Qbfplan.

**Bomb in Multiple Toilets.** The next domain, called BMTC($p,t$), is the generalization of the BTC problem to the case of multiple toilets ($p$ is the number of packages, while $t$ is the number of toilets). The problem becomes more parallelizable when the number of toilets increases. Furthermore, we considered three versions of the problem with increasing uncertainty in the initial states. In the first class of tests ("Low Uncertainty" columns), the only uncertainty is the position of the bomb which is unknown, while toilets are known to be not clogged. The "Mid Uncertainty" and "High Uncertainty" columns show the results in presence of more uncertainty in the initial state. In the second [third, respectively] class of tests, the status of every odd [every, resp.] toilet can be either clogged or not clogged. This increases the number of possible initial states.

The results are reported in Table 3 (for the comparison with Cgp) and in Table 4 (for the comparison with Gpt). The IS column represents the number of initial states of the corresponding problem. Cgp is able to fully exploit the parallelism of the problem. However, Cgp is never able to explore more than 9 levels in the planning graph, with depth decreasing with the number of initial states. The results also show that Cmbp and Gpt are much less sensitive to the number of initial states than Cgp. With increasing initial

| BMTC | | | Low Uncertainty | | | | Mid Uncertainty | | | | | High Uncertainty | | | | |
|---|---|---|---|---|---|---|---|---|---|---|---|---|---|---|---|---|
| | | | $C_{MBP}$ | | $C_{GP}$ | | | $C_{MBP}$ | | $C_{GP}$ | | | $C_{MBP}$ | | $C_{GP}$ | |
| (p,t) | IS | \|P\| | #BS / #BSH | Time | \|L\| | Time | IS | #BS / #BSH | Time | \|L\| | Time | IS | #BS / #BSH | Time | \|L\| | Time |
| (2,2) | 2 | 2 | 10 / 18 | 0.000 | 1 | 0.000 | 4 | 12 / 34 | 0.000 | 2 | 0.010 | 8 | 12 / 40 | 0.000 | 2 | 0.030 |
| (3,2) | 3 | 4 | 26 / 84 | 0.000 | 3 | 0.020 | 6 | 28 / 106 | 0.000 | 3 | 0.040 | 12 | 28 / 112 | 0.010 | 4 | 13.560 |
| (4,2) | 4 | 6 | 58 / 250 | 0.000 | 3 | 0.030 | 8 | 60 / 286 | 0.020 | 4 | 0.460 | 16 | 60 / 294 | 0.010 | 4 | 145.830 |
| (5,2) | 5 | 8 | 122 / 652 | 0.030 | 5 | 1.390 | 10 | 124 / 702 | 0.030 | 5 | 13,180 | 20 | 124 / 710 | 0.040 | 4 | — |
| (6,2) | 6 | 10 | 250 / 1552 | 0.070 | 5 | 3.490 | 12 | 252 / 1614 | 0.080 | 5 | — | 24 | 252 / 1622 | 0.080 | | |
| (7,2) | 7 | 12 | 506 / 3586 | 0.180 | 7 | 508.510 | 14 | 508 / 3662 | 0.190 | | | 28 | 508 / 3670 | 0.190 | | |
| (8,2) | 8 | 14 | 1018 / 8262 | 0.400 | 7 | 918.960 | 16 | 1020 / 8362 | 0.430 | | | 32 | 1020 / 8372 | 0.450 | | |
| (9,2) | 9 | 16 | 2042 / 18484 | 0.940 | 7 | — | 18 | 2044 / 18602 | 0.960 | | | 36 | 2044 / 18612 | 0.950 | | |
| (10,2) | 10 | 18 | 4090 / 40676 | 1.820 | | | 20 | 4092 / 40810 | 1.990 | | | 40 | 4092 / 40820 | 2.030 | | |
| (2,3) | 2 | 2 | 18 / 42 | 0.000 | 1 | 0.010 | 8 | 24 / 99 | 0.000 | 2 | 0.090 | 16 | 24 / 126 | 0.010 | 2 | 0.170 |
| (3,3) | 3 | 3 | 47 / 202 | 0.010 | 1 | 0.010 | 12 | 56 / 349 | 0.020 | 2 | 0.200 | 24 | 56 / 373 | 0.020 | 2 | 0.690 |
| (4,3) | 4 | 5 | 110 / 736 | 0.030 | 3 | 0.110 | 16 | 120 / 942 | 0.040 | 3 | 0.990 | 32 | 120 / 972 | 0.040 | 3 | — |
| (5,3) | 5 | 7 | 237 / 2034 | 0.080 | 3 | 0.170 | 20 | 248 / 2335 | 0.110 | 3 | — | 40 | 248 / 2371 | 0.120 | | |
| (6,3) | 6 | 9 | 492 / 5106 | 0.230 | 3 | 0.340 | 24 | 504 / 5520 | 0.250 | | | 48 | 504 / 5562 | 0.240 | | |
| (7,3) | 7 | 11 | 1003 / 12128 | 0.560 | 5 | 6248.010 | 28 | 101 / 12673 | 0.590 | | | 56 | 1016 / 12721 | 0.640 | | |
| (8,3) | 8 | 13 | 2026 / 27836 | 1.300 | 4 | — | 32 | 204 / 28530 | 1.350 | | | 64 | 2040 / 28584 | 1.330 | | |
| (9,3) | 9 | 15 | 4073 / 62470 | 3.330 | | | 36 | 408 / 63331 | 3.370 | | | 72 | 4088 / 63391 | 3.390 | | |
| (10,3) | 10 | 17 | 8168 / 138046 | 7.280 | | | 40 | 818 / 139092 | 7.460 | | | 80 | 8184 / 139158 | 7.430 | | |
| (2,4) | 2 | 2 | 29 / 75 | 0.010 | 1 | 0.000 | 8 | 29 / 75 | 0.000 | 1 | 0.020 | 32 | 48 / 332 | 0.020 | 2 | 1.610 |
| (3,4) | 3 | 3 | 92 / 492 | 0.020 | 1 | 0.010 | 12 | 108 / 808 | 0.030 | 2 | 0.290 | 48 | 112 / 960 | 0.040 | 2 | 8.690 |
| (4,4) | 4 | 4 | 206 / 1686 | 0.060 | 1 | 0.010 | 16 | 236 / 2356 | 0.080 | 2 | 0.730 | 64 | 240 / 2532 | 0.090 | 2 | 32.190 |
| (5,4) | 5 | 6 | 457 / 4987 | 0.190 | 3 | 0.500 | 20 | 492 / 5888 | 0.230 | 2 | — | 80 | 496 / 6092 | 0.240 | 3 | — |
| (6,4) | 6 | 8 | 964 / 12456 | 0.410 | 3 | 1.160 | 24 | 1004 / 13648 | 0.470 | | | 96 | 1008 / 13876 | 0.470 | | |
| (7,4) | 7 | 10 | 1983 / 29453 | 1.040 | 3 | 2.410 | 28 | 2028 / 31004 | 1.120 | | | 112 | 2032 / 31260 | 1.160 | | |
| (8,4) | 8 | 12 | 4026 / 68466 | 2.740 | 3 | 8.540 | 32 | 4076 / 70584 | 2.870 | | | 128 | 4080 / 70912 | 2.910 | | |
| (9,4) | 9 | 14 | 8117 / 153895 | 6.690 | 4 | — | 36 | 8172 / 15654 | 6.900 | | | 144 | 8176 / 156904 | 6.970 | | |
| (10,4) | 10 | 16 | 16304 / 339160 | 14.420 | | | 40 | 16364 / 34234 | 14.630 | | | 160 | 16368 / 342736 | 14.770 | | |
| (2,5) | 2 | 2 | 43 / 117 | 0.010 | 1 | 0.010 | 16 | 43 / 117 | 0.010 | 1 | 0.130 | 64 | 93 / 751 | 0.030 | 2 | 21.120 |
| (3,5) | 3 | 3 | 164 / 1031 | 0.040 | 1 | 0.020 | 24 | 212 / 2008 | 0.080 | 2 | 3.540 | 96 | 224 / 2591 | 0.120 | 2 | 138.430 |
| (4,5) | 4 | 4 | 416 / 4304 | 0.150 | 1 | 0.020 | 32 | 475 / 6375 | 0.260 | 2 | 6.320 | 128 | 480 / 6740 | 0.260 | 2 | 551.210 |
| (5,5) | 5 | 5 | 872 / 11763 | 0.490 | 1 | 0.050 | 40 | 987 / 15928 | 0.700 | 2 | 37.959 | 160 | 992 / 16393 | 0.730 | 2 | 1523.840 |
| (6,5) | 6 | 7 | 1875 / 31695 | 1.300 | 3 | 5.920 | 48 | 2011 / 37759 | 1.890 | 2 | — | 192 | 2016 / 38334 | 1.980 | 2 | — |
| (7,5) | 7 | 9 | 3901 / 78009 | 3.990 | 3 | 18.410 | 56 | 4059 / 86716 | 4.480 | | | 224 | 4064 / 87411 | 4.540 | | |
| (8,5) | 8 | 11 | 7974 / 183036 | 9.670 | 3 | 62.040 | 64 | 8157 / 195055 | 10.590 | | | 256 | 8160 / 195880 | 10.640 | | |
| (9,5) | 9 | 13 | 16142 / 416333 | 24.250 | 3 | 194.640 | 72 | 16347 / 432408 | 25.600 | | | 288 | 16352 / 433373 | 25.370 | | |
| (10,5) | 10 | 15 | 32501 / 927329 | 54.910 | 3 | 289.680 | 80 | 32731 / 948279 | 56.420 | | | 320 | 32736 / 949394 | 56.290 | | |
| (2,6) | 2 | 2 | 60 / 168 | 0.010 | 1 | 0.010 | 16 | 60 / 168 | 0.010 | 1 | 0.200 | 128 | 171 / 1533 | 0.040 | 2 | 337.604 |
| (3,6) | 3 | 3 | 270 / 1848 | 0.070 | 1 | 0.010 | 24 | 270 / 1848 | 0.070 | 1 | 0.830 | 192 | 448 / 6248 | 0.310 | 2 | 1459.110 |
| (4,6) | 4 | 4 | 786 / 9294 | 0.300 | 1 | 0.040 | 32 | 920 / 13810 | 0.500 | 2 | 30.630 | 256 | 960 / 16344 | 0.690 | 2 | 5643.450 |
| (5,6) | 5 | 5 | 1777 / 29075 | 1.160 | 1 | 0.060 | 40 | 1958 / 37636 | 1.940 | 2 | 30.140 | 320 | 1984 / 39710 | 2.120 | 2 | — |
| (6,6) | 6 | 6 | 3613 / 71123 | 3.290 | 1 | 0.040 | 48 | 4005 / 90111 | 4.080 | 2 | 57.300 | 384 | 4032 / 92772 | 4.600 | | |
| (7,6) | 7 | 8 | 7625 / 180127 | 9.060 | 3 | 211.720 | 56 | 8100 / 208050 | 10.130 | 2 | — | 448 | 8128 / 211370 | 10.400 | | |
| (8,6) | 8 | 10 | 15726 / 429198 | 20.710 | 3 | 1015.160 | 64 | 16291 / 469277 | 22.620 | | | 512 | 16320 / 473328 | 23.000 | | |
| (9,6) | 9 | 12 | 32012 / 986188 | 50.610 | 3 | 3051.990 | 72 | 32674 / 1.04173e+06 | 53.510 | | | 576 | 32704 / 1.04658e+06 | 54.010 | | |
| (10,6) | 10 | 14 | 64675 / 2.21106e+06 | 111.830 | 2 | — | 80 | 65441 / 2.28585e+06 | 116.440 | | | 640 | 65472 / 2.29158e+06 | 116.240 | | |

Table 3: Results for the BMTC problems.

| BMTC | Low Unc. | | High Unc. | |
|---|---|---|---|---|
| | CMBP | GPT | CMBP | GPT |
| (p,t) | Time | Time | Time | Time |
| (2,2) | 0.000 | 0.079 | 0.010 | 0.079 |
| (3,2) | 0.010 | 0.087 | 0.010 | 0.091 |
| (4,2) | 0.000 | 0.105 | 0.020 | 0.121 |
| (5,2) | 0.040 | 0.146 | 0.040 | 0.198 |
| (6,2) | 0.080 | 0.227 | 0.070 | 0.376 |
| (7,2) | 0.190 | 0.441 | 0.200 | 0.850 |
| (8,2) | 0.390 | 0.922 | 0.400 | 1.966 |
| (9,2) | 0.910 | 2.211 | 0.950 | 4.743 |
| (10,2) | 1.850 | 5.169 | 1.900 | 10.620 |
| (2,4) | 0.000 | 0.109 | 0.010 | 0.121 |
| (3,4) | 0.010 | 0.156 | 0.040 | 0.284 |
| (4,4) | 0.050 | 0.270 | 0.100 | 1.016 |
| (5,4) | 0.180 | 0.616 | 0.240 | 3.282 |
| (6,4) | 0.370 | 1.435 | 0.460 | 9.374 |
| (7,4) | 1.080 | 3.484 | 1.190 | 27.348 |
| (8,4) | 2.700 | 8.767 | 2.830 | 72.344 |
| (9,4) | 8.970 | 23.858 | 6.920 | 180.039 |
| (10,4) | 14.210 | 59.966 | 114.690 | 440.308 |
| (2,6) | 0.010 | 0.303 | 0.060 | 0.482 |
| (3,6) | 0.050 | 0.562 | 0.260 | 2.471 |
| (4,6) | 0.310 | 1.354 | 0.620 | 17.406 |
| (5,6) | 1.110 | 3.257 | 2.060 | 74.623 |
| (6,6) | 3.400 | 8.691 | 4.660 | 243.113 |
| (7,6) | 8.910 | 25.677 | 10.430 | 701.431 |
| (8,6) | 21.240 | 68.427 | 23.860 | = |
| (9,6) | 49.880 | 289.000 | 54.190 | |
| (10,6) | 113.680 | 486.969 | 118.590 | |

Table 4: Results for the BMTC problems.

uncertainty, CGP is almost unable to solve what were trivial problems. GPT performs better than CGP, but it suffers from the explicit representation of the search space.

**Bomb in the Toilet with Uncertain Clogging.** The BTUC($p$) domain is the domain described in Section 2, where clogging is an uncertain outcome of dunking a package. This kind of problem cannot be expressed in CGP. The results for CMBP and GPT are reported in Table 5. Although CMBP performs better than GPT (by a factor of two to three), there is no significant difference in the behavior. It is interesting to compare the results of CMBP for the BTC and BTUC problems. For GPT a slight difference is noticeable, resulting from the increased branching factor in the search space due to the uncertainties in the effects of action executions. In the performance of CMBP, the number of uncertainties is not a direct factor — for example, in the BTC(16) and BTUC(16), the performance is almost the same.

### 7.2.2 RING OF ROOMS

**Simple Ring of Room.** We considered another domain, where a robot can move in a ring of rooms. Each room has a window, which can be either open, closed or locked. The robot can move (either clockwise or counterclockwise), close the window of the room where it is, and lock it if closed. The goal is to have all windows locked.

|  | \|P\| | #BS/#BSH | Time | Time |
|---|---|---|---|---|
|  |  | CMBP |  | GPT |
| BTUC(2) | 3 | 6 / 8 | 0.000 | 0.076 |
| BTUC(3) | 5 | 14 / 23 | 0.000 | 0.078 |
| BTUC(4) | 7 | 30 / 61 | 0.010 | 0.085 |
| BTUC(5) | 9 | 62 / 150 | 0.010 | 0.098 |
| BTUC(6) | 11 | 126 / 347 | 0.030 | 0.128 |
| BTUC(7) | 13 | 254 / 796 | 0.050 | 0.205 |
| BTUC(8) | 15 | 510 / 1844 | 0.170 | 0.380 |
| BTUC(9) | 17 | 1022 / 4149 | 0.310 | 0.812 |
| BTUC(10) | 19 | 2046 / 9190 | 0.720 | 1.828 |
| BTUC(16) | 31 | 131070 / 921355 | 98.270 | 486.252 |

Table 5: Results for the BTUC problems.

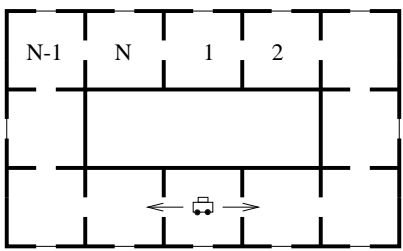

In the problem RING($r$), where $r$ is the number of rooms, the uncertainty is only in the initial condition: both the position of the robot and the status of the windows can be uncertain. These problems do not have a parallel solution, and have a large number of initial states ($r * 3^r$), corresponding to full uncertainty on the position of the robot and on the status of each window. The results[8] are reported on the left in Table 6. CMBP outperforms both CGP and GPT, although GPT performs much better than CGP. Both CGP and GPT suffer from the increasing complexity of the problem. On the right in Table 6, we plot (for the RING(5) problem) the dependency of CGP on the number of initial states combined with the number of levels to be explored (different goals were provided which require the exploration of different levels). It is clear that the number of initial states and the depth of the search are both critical factors for CGP.

|  | \|P\| | #BS/#BSH | Time | \|L\| | Time | Time |
|---|---|---|---|---|---|---|
|  |  | CMBP |  |  | CGP | GPT |
| RING(2) | 5 | 8 / 24 | 0.000 | 3 | 0.070 | 0.085 |
| RING(3) | 8 | 26 / 78 | 0.020 | 4 | — | 0.087 |
| RING(4) | 11 | 80 / 240 | 0.040 |  |  | 0.392 |
| RING(5) | 14 | 242 / 726 | 0.120 |  |  | 1.150 |
| RING(6) | 17 | 728 / 2184 | 0.370 |  |  | 6.620 |
| RING(7) | 20 | 2186 / 6558 | 1.420 |  |  | 23.636 |
| RING(8) | 23 | 6560 / 19680 | 4.950 |  |  | 105.158 |
| RING(9) | 26 | 19682 / 59046 | 27.330 |  |  | == |
| RING(10) | 29 | 59048 / 177144 | 106.870 |  |  |  |

| CGP on RING(5) | | | | |
|---|---|---|---|---|
| IS | \|L\| | Time | \|L\| | Time |
| 1 | 5 | 0.010 | 9 | 0.020 |
| 2 | 5 | 0.060 | 9 | 0.140 |
| 4 | 5 | 0.420 | 9 | 1.950 |
| 8 | 5 | 6.150 | 9 | 359.680 |
| 16 | 5 | — | 9 | — |

Table 6: The results for the RING problems.

---

8. The times reported for CGP refer to a scaled-down version of the problem, where locking is not taken into account, and thus the maximum number of initial states is $r * 2^r$.

**Ring of Rooms with Uncertain Action Effects.** We considered a variation of the RING domain, called URING, first introduced by Cimatti and Roveri (1999), which is not expressible in CGP. If a window is not locked and the robot is not performing an action which will determine its status (e.g. closing it), then the window can open or close nondeterministically. For instance, while the robot is moving from room 1 to room 2, the windows in room 3 and 4 could be open or closed by the wind. This domain is clearly designed to stress the ability of a planner to deal with actions having a large number of resulting states. In the worst case (e.g. a move action performed when no window is locked), there are $2^r$ possible resulting states. Although seemingly artificial, this captures the fact that environments can be in practice highly nondeterministic. We tried to compare CMBP and GPT on the URING problem. In principle GPT is able to deal with uncertainty in the action effects. However, we failed to codify the URING in the GPT language, because it requires a conditional description of uncertain effects. Therefore, we experimented with a variation of the RING domain featuring a higher degree of nondeterminism, called NDRING in the following. The NDRING domain contains an increasing number of additional propositions, called in the following *noninertial* propositions, which are initially unknown and are nondeterministically altered by each action. If $i$ is the number of noninertial propositions, each action has $2^i$ possible outcomes. The results are listed in Table 7, with columns labeled with Time($i$). The growing branching factor during the search has a major impact on the performance of GPT, while CMBP is insensitive to this kind of uncertainty. (The performance of CMBP for a lower number of noninertial propositions are not reported because they are basically the same.)

| | |P| | #BS/#BSH | Time (5) | Time (2) | Time (3) | Time (4) | Time (5) |
|---|---|---|---|---|---|---|---|
| | | CMBP | | | GPT | | | |
| NDRING(2) | 5 | 8 / 24 | 0.000 | 0.140 | 0.384 | 0.948 | 4.544 |
| NDRING(3) | 8 | 26 / 78 | 0.020 | 0.256 | 0.679 | 2.574 | 13.960 |
| NDRING(4) | 11 | 80 / 240 | 0.040 | 1.046 | 3.025 | 12.548 | 67.714 |
| NDRING(5) | 14 | 242 / 726 | 0.110 | 4.550 | 12.960 | 48.426 | = |
| NDRING(6) | 17 | 728 / 2184 | 0.350 | 18.758 | 57.300 | = | |
| NDRING(7) | 20 | 2186 / 6558 | 1.350 | 108.854 | = | | |
| NDRING(8) | 23 | 6560 / 19680 | 4.990 | = | | | |
| NDRING(9) | 26 | 19682 / 59046 | 27.060 | | | | |
| NDRING(10) | 29 | 59048 / 177144 | 103.760 | | | | |

Table 7: The results for the NDRING problems.

The URING problem was run only on CMBP. The results are listed in Table 8. It can be noticed that the performances of CMBP improve significantly with respect to the RING problem. This can be explained considering that, despite the larger number of transitions, the number of explored belief states is significantly smaller (see the Bs cache statistics in Tables 6 and 8).

### 7.2.3 SQUARE AND CUBE

The following domains are the SQUARE($n$) and CUBE($n$) from the GPT distribution (Bonet & Geffner, 2000). These problems consist of a robot navigating in a square or cube of side $n$. In both domains there are actions for moving the robot in all the possible directions. Moving the robot against a boundary leaves the robot in the same position. The original problems, called CORNER in the following, require the robot to reach a corner, starting from a completely unspecified position. We introduced two variations. In the first, called FACE, the initial position is any position of a given side [face] of the square [cube], while the goal is to reach the central position of the opposite side [face]. In the second, called CENTER, the initial position is completely unspecified, and the goal is the center of the square [cube]. For the corner problem, a simple heuristic is to perform only steps towards the corner, thus pruning half of the actions. The variations are designed not to allow for a simple heuristic — for instance, in the CENTER problem, no action can be eliminated.

| | | Cmbp | |
|---|---|---|---|
| | \|P\| | #BS/#BSH | Time |
| URING(2) | 5 | 5 / 16 | 0.000 |
| URING(3) | 8 | 11 / 34 | 0.010 |
| URING(4) | 11 | 23 / 70 | 0.020 |
| URING(5) | 14 | 47 / 142 | 0.040 |
| URING(6) | 17 | 95 / 286 | 0.080 |
| URING(7) | 20 | 191 / 574 | 0.190 |
| URING(8) | 23 | 383 / 1150 | 0.410 |
| URING(9) | 26 | 767 / 2302 | 0.980 |
| URING(10) | 29 | 1535 / 4606 | 2.2300 |

Table 8: Results for the URING problems.

| SQUARE(i) | CORNER | | | | FACE | | | | CENTER | | | |
|---|---|---|---|---|---|---|---|---|---|---|---|---|
| | Cmbp | | | Gpt | Cmbp | | | Gpt | Cmbp | | | Gpt |
| | \|P\| | #BS/#BSH | Time | Time | \|P\| | #BS/#BSH | Time | Time | \|P\| | #BS/#BSH | Time | Time |
| SQUARE(2) | 2 | 2 / 4 | 0.000 | 0.074 | 2 | 2 / 4 | 0.000 | 0.058 | 2 | 2 / 4 | 0.000 | 0.060 |
| SQUARE(4) | 6 | 15 / 37 | 0.000 | 0.080 | 7 | 33 / 83 | 0.000 | 0.065 | 8 | 76 / 190 | 0.010 | 0.083 |
| SQUARE(6) | 10 | 35 / 93 | 0.000 | 0.092 | 12 | 86 / 232 | 0.020 | 0.089 | 14 | 218 / 592 | 0.040 | 0.216 |
| SQUARE(8) | 14 | 63 / 173 | 0.020 | 0.115 | 17 | 163 / 453 | 0.040 | 0.139 | 20 | 432 / 1210 | 0.090 | 0.695 |
| SQUARE(10) | 18 | 99 / 277 | 0.030 | 0.149 | 22 | 264 / 746 | 0.090 | 0.228 | 26 | 718 / 2044 | 0.190 | 2.135 |
| SQUARE(12) | 22 | 143 / 405 | 0.050 | 0.196 | 27 | 389 / 1111 | 0.150 | 0.371 | 32 | 1076 / 3094 | 0.360 | 5.340 |
| SQUARE(14) | 26 | 195 / 557 | 0.070 | 0.261 | 32 | 538 / 1548 | 0.230 | 0.582 | 38 | 1506 / 4360 | 0.560 | 12.284 |
| SQUARE(16) | 30 | 255 / 733 | 0.080 | 0.357 | 37 | 711 / 2057 | 0.320 | 0.908 | 44 | 2008 / 5842 | 0.820 | 26.241 |
| SQUARE(18) | 34 | 323 / 933 | 0.120 | 0.503 | 42 | 908 / 2638 | 0.540 | 1.343 | 50 | 2582 / 7540 | 1.330 | 52.091 |
| SQUARE(20) | 38 | 399 / 1157 | 0.160 | 0.638 | 47 | 1129 / 3291 | 0.650 | 1.883 | 56 | 3228 / 9454 | 1.790 | 94.204 |

| CUBE(i) | CORNER | | | | FACE | | | | CENTER | | | |
|---|---|---|---|---|---|---|---|---|---|---|---|---|
| | Cmbp | | | Gpt | Cmbp | | | Gpt | Cmbp | | | Gpt |
| | \|P\| | #BS/#BSH | Time | Time | \|P\| | #BS/#BSH | Time | Time | \|P\| | #BS/#BSH | Time | Time |
| CUBE(2) | 3 | 6 / 19 | 0.000 | 0.332 | 3 | 6 / 19 | 0.000 | 0.061 | 3 | 6 / 19 | 0.010 | 0.061 |
| CUBE(3) | 6 | 26 / 99 | 0.010 | 0.168 | 6 | 26 / 99 | 0.000 | 0.069 | 6 | 26 / 99 | 0.010 | 0.144 |
| CUBE(4) | 9 | 63 / 261 | 0.020 | 0.430 | 11 | 319 / 1360 | 0.050 | 0.193 | 12 | 722 / 3091 | 0.130 | 0.569 |
| CUBE(5) | 12 | 124 / 537 | 0.040 | 0.276 | 14 | 709 / 3095 | 0.220 | 0.412 | 15 | 1696 / 7402 | 0.430 | 2.010 |
| CUBE(6) | 15 | 215 / 957 | 0.050 | 0.500 | 19 | 1343 / 6116 | 0.430 | 1.479 | 21 | 3365 / 15432 | 0.910 | 10.717 |
| CUBE(7) | 18 | 342 / 1551 | 0.100 | 0.567 | 22 | 2255 / 10377 | 0.840 | 3.323 | 24 | 5797 / 26814 | 1.860 | 34.074 |
| CUBE(8) | 21 | 511 / 2349 | 0.160 | 1.082 | 27 | 3519 / 16464 | 1.400 | 8.161 | 30 | 9248 / 43541 | 3.520 | 109.852 |
| CUBE(9) | 24 | 728 / 3381 | 0.330 | 1.765 | 30 | 5169 / 24331 | 2.810 | 16.272 | 33 | 13786 / 65237 | 7.260 | 701.910 |
| CUBE(10) | 27 | 999 / 4677 | 0.440 | 2.068 | 35 | 7279 / 34564 | 4.550 | 32.226 | 39 | 19667 / 93898 | 9.990 | = |
| CUBE(15) | 42 | 3374 / 16167 | 1.940 | 9.207 | 54 | 26439 / 127825 | 28.560 | = | 60 | 74041 / 359354 | 58.930 | |

Table 9: Results for the SQUARE and CUBE problems.

The results for these problems are reported in Table 9. The tests were run only with Cmbp and Gpt. The experiments highlight that the efficiency of Gpt strongly depends on the quality of the heuristic function. If, as in the first set of experiments, the heuristics are effective, then GPT is almost as good as CMBP. Otherwise, GPT degrades significantly. In general, finding heuristics which are effective in the belief space appears to be a nontrivial problem. CMBP appears to be more stable[9], as it performs a blind, breadth-first search, and relies on the cleverness of the symbolic representation to achieve efficiency.

### 7.2.4 OMELETTE

Finally, we considered the OMELETTE($i$) problem (Levesque, 1996). The goal is to have $i$ good eggs and no bad ones in one of two bowls of capacity $i$. There is an unlimited number of eggs, each of which can be unpredictably good or bad. The eggs can be grabbed and broken into a bowl. The content of a bowl can be discarded, or poured to the other bowl. Breaking a rotten egg in a bowl has the effect of spoiling the bowl. A bowl can always be cleaned by discarding its content. The problem is originally presented as a partial observability problem, with a sensing action allowing to test if a bowl is spoiled or not. We considered the variation of the problem without sensing action: in this case no conformant solution exists. We used the OMELETTE problems to test the ability of CMBP and GPT to discover that the problem admits no conformant solution. The results are reported in Table 10. The table shows that CMBP is very effective in checking the absence of a conformant solution, and outperforms GPT by several orders of magnitude.

| | CMBP | | | GPT |
|---|---|---|---|---|
| | # steps | #BS/#BSH | Time | Time |
| OMELETTE(3) | 9 | 15 / 34 | 0.020 | 0.237 |
| OMELETTE(4) | 11 | 19 / 42 | 0.030 | 0.582 |
| OMELETTE(5) | 13 | 23 / 50 | 0.040 | 1.418 |
| OMELETTE(6) | 15 | 27 / 58 | 0.050 | 2.904 |
| OMELETTE(7) | 17 | 31 / 66 | 0.060 | 5.189 |
| OMELETTE(8) | 19 | 35 / 74 | 0.090 | 10.307 |
| OMELETTE(9) | 21 | 39 / 82 | 0.110 | 18.744 |
| OMELETTE(10) | 23 | 43 / 90 | 0.120 | 32.623 |
| OMELETTE(15) | 33 | 63 / 130 | 0.210 | 225.530 |
| OMELETTE(20) | 43 | 83 / 170 | 0.440 | == |
| OMELETTE(30) | 63 | 123 / 250 | 0.890 | |

Table 10: Results for the OMELETTE problems.

## 7.3 Summarizing Remarks

Overall, CMBP appears to implement the most effective approach to conformant planning, both in terms of expressivity and performance. CGP is only able to deal with uncertainties in the initial states, and cannot conclude that the problem does not admit a conformant solution. The main problem in CGP seems to be its enumerative approach to uncertainties, and the increased number of initial states severely affects the performance (see Table 3 and Table 6).

QBFPLAN is in principle able to deal with uncertain action effects, but cannot conclude that the problem does not admit a conformant solution. From the small number of ex-periments that we could perform, the approach implemented by QBFPLAN is limited by the SATPLAN style of search: the intermediate results obtained while solving an encoding at depth $k$ are not reused while solving encodings of increasing depth. Furthermore, the solver appears to be specialized in finding a model, rather than in proving unsatisfiability. However, the latter ability is needed in all encodings but the final one.

---

9. Consider also that the problems are increasingly more difficult (see for instance the plan length).

GPT is a very expressive system, which allows efficiently dealing with a wide class of planning problems. As far as conformant planning is concerned, it is as expressive as CMBP. It allows dealing with uncertain action effects, and can conclude that a problem does not have a conformant solution. However, CMBP appears to outperform GPT in several respects. First, the behaviour of GPT appears to be directly related to the number of possible outcomes in an action. Furthermore, the efficiency of GPT depends on the effectiveness of the heuristic functions, which can be sometimes difficult to devise, and cannot help when the problem does not admit a solution.

The main strength of CMBP is its independence on the *number* of uncertainties, which is achieved with the use of symbolic techniques. Being fully symbolic, CMBP does not exhibit the enumerative behaviour of its competitors. Compared to the original approach described by Cimatti and Roveri (1999), a substantial improvement of the performance has been obtained by the new implementation of the pruning step. A disclaimer is in order. It is well known that BDD based computations are subject to a blow-up in memory requirements when computing certain classes of boolean functions, e.g. multipliers (Bryant, 1986). It would be trivial to make up an example where the performance of CMBP degrades exponentially. However, in none of the examples we considered, which included all the examples in the distribution of CGP and GPT, this phenomenon occurred.

## 8. Other Related Work

The term *conformant planning* was first introduced by Goldman (1996), while presenting a formalism for constructing conformant plans based on an extension of dynamic logic. Recently, Ferraris and Giunchiglia (2000) presented another conformant planner based on SAT techniques. The system is not available for a direct comparison with CMBP. The effectiveness of the approach is difficult to evaluate, as only a limited testing is described (Ferraris & Giunchiglia, 2000). The performance is claimed to be comparable with CGP. However, the results are reported only for the encoding corresponding to the solution, and the behaviour of QBFPLAN reported in Table 2 suggests that this kind of analysis might be limited.

Several works share the idea of planning based on automata theory. The most closely related are the works in the lines of planning via model checking (Cimatti et al., 1997), upon which our work is based. This approach allows, for instance, to automatically construct universal plans which are guaranteed to achieve the goal in a finite number of steps (Cimatti et al., 1998b), or which implement trial-and-error strategies (Cimatti et al., 1998a; Daniele et al., 1999). These results are obtained under the hypothesis of total observability, while here run-time observation is not available. The main difference is that a substantial extension is required to lift symbolic techniques to search in the space of belief states. De Giacomo and Vardi (1999) analyze several forms of planning in the automata theoretic framework. Goldman, Musliner and Pelican (2000) present a method where model checking in timed automata is interleaved with the plan formation activity, to make sure that the timing constraints are met. Finally, Hoey and his colleagues (1999) use algebraic decision diagrams to tackle the problem of stochastic planning.

## 9. Conclusions and Future Work

In this paper we presented a new approach to conformant planning, based on the use of Symbolic Model Checking techniques. The algorithm is very general, and applies to complex planning domains, with uncertainty in the initial condition and in action effects, which can be described as finite state automata. The algorithm is based on a breadth-first, backward search, and returns conformant plans of minimal length, if a solution to the planning problem exists. Otherwise, it terminates with failure. The algorithm is designed to take full advantage of the symbolic representation based on BDDs. The implementation of the approach in the CMBP system has been highly optimized, in particular in the crucial step of termination checking. We performed an experimental comparison of our approach with the state of the art conformant planners CGP, QBFPLAN and GPT. CMBP is strictly more expressive than QBFPLAN and CGP. On all the problems for which a comparison was possible, CMBP outperformed its competitors in terms of run times, sometimes by orders of magnitude. Thanks to the use of symbolic data structures, CMBP is able to deal efficiently with problems with large numbers of initial states and action outcomes. On the other hand, the qualitative behavior of CGP and GPT seems to depend heavily on the enumerative nature of their algorithms. Differently from GPT, CMBP is independent of the effectiveness of the heuristic used to drive the search.

The research presented in this paper will be extended in the following directions. First, we are investigating an alternative approach to conformant planning, where the breadth-first style of the search is given up. These techniques appear to be extremely promising — preliminary experiments have led to speed ups of up to two orders of magnitude over the results presented in this paper for problems which admit a solution. Second, we will tackle the problem of conditional planning under partial observability, under the hypothesis that a limited amount of information can be acquired at run time. As conformant planning, this problem can be seen as search in the belief space. However, it appears to be significantly complicated by the need for dealing with run-time observation and conditional plans. Finally, we are considering the extension of the domain construction of the planner with more expressive input language, such as $\mathcal{C}$, and invariant detection techniques.

## Acknowledgements

Fausto Giunchiglia provided continuous encouragement and feedback on this work. We thank Piergiorgio Bertoli, Blai Bonet, Marco Daniele, Hector Geffner, Enrico Giunchiglia, Jussi Rintanen, David Smith, Paolo Traverso, Dan Weld for valuable discussions on conformant planning and various comments on this paper. David Smith provided the code of CGP, a large number of examples, and the time-out mechanism used in the experimental evaluation. Jussi Rintanen made QBFPLAN available under Linux.